%% file: main.tex
\documentclass[10pt,twocolumn,letterpaper]{article}

\usepackage{cvpr}
\usepackage{times}
\usepackage{epsfig}
\usepackage{graphicx}
\usepackage{amsmath}
\usepackage{amssymb}
\usepackage{xcolor}
\usepackage{makecell}

\usepackage[pagebackref=true,breaklinks=true,letterpaper=true,colorlinks,bookmarks=false]{hyperref}

\cvprfinalcopy % *** Uncomment this line for the final submission

\def\cvprPaperID{177} % *** Enter the CVPR Paper ID here

\newcommand{\comment}[1]{}

\newcommand{\PARTS}{PartNet}

\usepackage{textcomp}
\newcommand{\textapprox}{\raisebox{0.15ex}{\texttildelow}}

\begin{document}

%%%%%%%%% TITLE
\title{PartNet: A Large-scale Benchmark for Fine-grained and \\Hierarchical Part-level 3D Object Understanding}

\author{
Kaichun Mo$^{1}$ \, Shilin Zhu$^{2}$  \, Angel X. Chang$^{3}$ \, Li Yi$^{1}$ \,
%\vspace{0.1cm} \\ 
Subarna Tripathi$^{4}$  \, Leonidas J. Guibas$^{1}$ \, Hao Su$^{2}$
\vspace{0.2cm} \\ 
$^{1}$Stanford University \, $^{2}$University of California San Diego \,  $^{3}$Simon Fraser University \, $^{4}$Intel AI Lab
\vspace{0.2cm} \\
\href{https://cs.stanford.edu/~kaichun/partnet/}{https://cs.stanford.edu/\textapprox kaichun/partnet/}
}

\maketitle
%\thispagestyle{empty}

%%%%%%%%% ABSTRACT
\label{abs}
\input{tex/abstract.tex}

%%%%%%%%% BODY TEXT
\section{Introduction}
\label{sec:intro}

\input{tex/intro.tex}

\section{Related Work}
\label{sec:related}
\input{tex/related.tex}

\section{Data Annotation}
\label{sec:framework}
\input{tex/framework.tex}

\section{PartNet Dataset}
\label{sec:data}

\input{tex/data.tex}

\section{Tasks and Benchmarks}
\label{sec:exp}
\input{tex/exp.tex}

\section{Conclusion}
\label{sec:conclusion}
\input{tex/conclusion.tex}

\section*{Acknowledgements}
\input{tex/ack.tex}

{\small
\bibliographystyle{ieee}
\bibliography{ref}
}

\appendix

\section{Overview}
This document provides additional dataset visualization and statistics (Sec~\ref{sec:data}), hierarchical template design details and visualization (Sec~\ref{sec:template}), and the architectures and training details for the three shape segmentation tasks (Sec~\ref{sec:task}), to the main paper. 

\section{More Dataset Visualization and Statistics}
\label{sec:data}
\input{tex_supp/data.tex}

\section{More Template Design Details and Visualization}
\label{sec:template}
\input{tex_supp/template.tex}

\section{Tasks and Benchmarks}
\label{sec:task}

\input{tex_supp/tasks.tex}

\input{tex_supp/figures-min.tex}

\end{document}

%% file: tex/abstract.tex
\begin{abstract}
%Detecting and reasoning about object parts in 3D is critical for creating robots that can interact with the world.  
We present \PARTS{}: a consistent, large-scale dataset of 3D objects annotated with fine-grained, instance-level, and hierarchical 3D part information. Our dataset consists of 573,585 part instances over 26,671 3D models covering 24 object categories. This dataset enables and serves as a catalyst for many tasks such as shape analysis, dynamic 3D scene modeling and simulation, affordance analysis, and others. Using our dataset, we establish three benchmarking tasks for evaluating 3D part recognition: fine-grained semantic segmentation, hierarchical semantic segmentation, and instance segmentation. We benchmark four state-of-the-art 3D deep learning algorithms for fine-grained semantic segmentation and three baseline methods for hierarchical semantic segmentation. We also propose a novel method for part instance segmentation and demonstrate its superior performance over existing methods.
\end{abstract}

%% file: tex/intro.tex
% Part knowledge is critical for understanding the structure and semantics of objects in the real world.
% People recognize, organize, and interact with objects base on the knowledge of their parts~\cite{hoffman1984parts}.
Being able to parse objects into parts is critical for humans to understand and interact with the world. People recognize, categorize, and organize objects based on the knowledge of their parts~\cite{hoffman1984parts}. Many actions that people take in the real world require detection of parts and reasoning over parts. For instance, we open doors using doorknobs and pull out drawers by grasping their handles.  Teaching machines to analyze parts is thus essential for many vision, graphics, and robotics applications, such as predicting object functionality~\cite{hu2016learning,hu2018predictive}, human-object interactions~\cite{kim2014shape2pose}, simulation~\cite{kolve2017ai2}, shape editing~\cite{ovsjanikov2011exploration,jain2012exploring}, and shape generation~\cite{li2017grass,wu2018structure}. % These applications necessitate an understanding of the \emph{hierarchical connectivity} of parts in a \emph{fine-grained} and \emph{instance-level} manner.

To enable part-level object understanding by learning approaches, 3D data with part annotations are in high demand. Many cutting-edge learning algorithms, especially for 3D understanding~\cite{yi2016scalable,yi2017learning,qi2017pointnet}, intuitive physics~\cite{liu2018physical}, and reinforcement learning~\cite{zhu2017visual,puig2018virtualhome}, require such data to train the networks and benchmark the performances. 
%\angel{following sentences aren't really supported by our dataset, big leap to state. maybe reword to say our part dataset is a step towards building large scale interactable objects and scenes.} 
Researchers are also increasingly interested in synthesizing dynamic data through physical simulation engines~\cite{kolve2017ai2,yan2018chalet,puig2018virtualhome}. Creation of large-scale animatable scenes will require a large amount of 3D data with affordances and mobility information. Object parts serve as a critical stepping stone to access this information.  Thus it is necessary to have a big 3D object dataset with part annotation. % which could be an extremely useful resource for many cutting-edge tasks, such as robotic learning \cite{zhu2017visual,puig2018virtualhome} and intuitive physics 

\begin{figure}[t]
\centering
   \includegraphics[width=\linewidth]{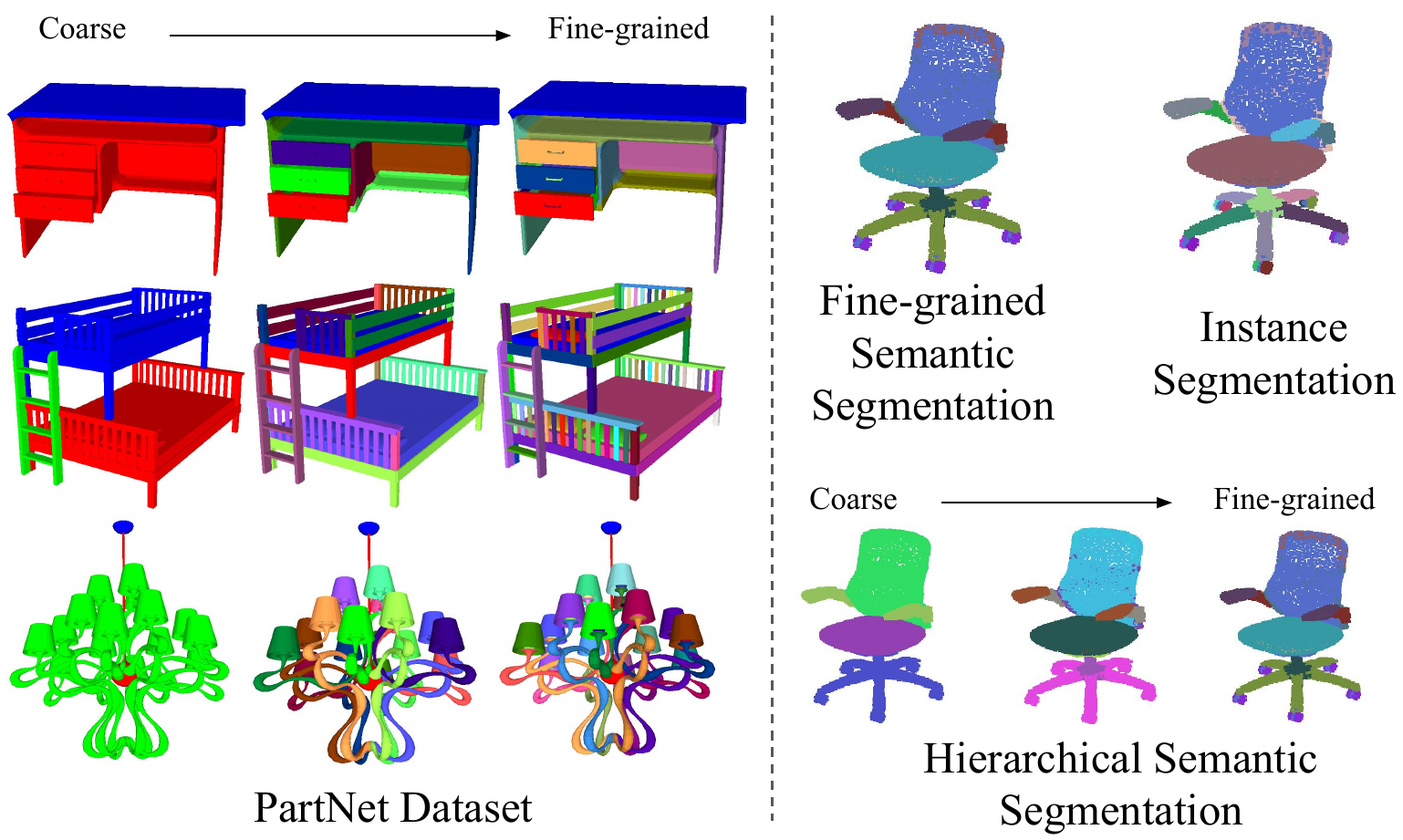}
   \caption{\textbf{PartNet dataset and three benchmarking tasks.} Left: we show example annotations at three levels of segmentation in the hierarchy. Right: we propose three fundamental and challenging segmentation tasks and establish benchmarks using PartNet.}
   \label{fig:teaser}
   \vspace{-3mm}
\end{figure}

\begin{table*}[t]
  \centering
  \small
  \begin{tabular}[width=\linewidth]{c|cccccccc}\Xhline{2\arrayrulewidth}
  \textbf{Dataset} &  \textbf{\#Shape}  & \textbf{\#Part}  &  \textbf{\#Category}  & \textbf{Granularity}     &  \textbf{Semantics}  & \textbf{Hierarchical}   & \textbf{Instance-level} &  \textbf{Consistent} \\ \Xhline{2\arrayrulewidth}
  \textbf{Chen et al.} \cite{chen2009benchmark} &  380 & 4,300  &  19    &   \textbf{Fine-grained}  & No & No  & \textbf{Yes} & No \\
  \textbf{MCL} \cite{wang2018learning} &   1,016 & 7,537 & 10   &  \textbf{Fine-grained} &  \textbf{Yes} &  No  &  No   &  \textbf{Yes}  \\
  \textbf{Chang et al.} \cite{chang2018linking} & 2,278 & 27,477  &  \textbf{90}  &  \textbf{Fine-grained}  & \textbf{Yes} &  No & \textbf{Yes} &  No \\
  \textbf{Yi et al.} \cite{yi2016scalable} &     \textbf{31,963} &  80,323    &   16     &  Coarse  &  \textbf{Yes} &  No  &  No  &  \textbf{Yes}  \\
  \hline
  \textbf{PartNet (ours)}    &   26,671  &  \textbf{573,585}  &  24   &     \textbf{Fine-grained}   & \textbf{Yes} &   \textbf{Yes}     &   \textbf{Yes}      & \textbf{Yes}         \\
    \Xhline{2\arrayrulewidth}
  \end{tabular}
  \vspace{-1mm}
  \caption{Comparison to the other shape part datasets.}
  \label{tab:datacompare}
  \vspace{-3mm}
\end{table*}

With the availability of the existing 3D shape datasets with part annotations~\cite{chen2009benchmark,chang2015shapenet,yi2016scalable}, we witness increasing research interests and advances in 3D part-level object understanding.
%increasing availability of 3D data and methods that analyze and process such data. 
Recently, a variety of learning methods have been proposed to push the state-of-the-art for 3D shape segmentation \cite{qi2017pointnet, qi2017pointnet++, yi2017syncspeccnn, klokov2017escape, wang2017cnn, li2018so, hermosilla2018monte, wang2018dynamic, wang2018voxsegnet, xu2018spidercnn, su2018splatnet, graham20183d, li2018pointcnn, le2018pointgrid}.
%Many of these works have looked into 3D segmentation of objects into constituent parts based on existing 3D datasets~\cite{chang2015shapenet}.
However, existing datasets only provide part annotations on relatively small numbers of object instances~\cite{chen2009benchmark}, or on coarse yet non-hierarchical part annotations~\cite{yi2016scalable}, restricting the applications that involves understanding fine-grained and hierarchical shape segmentation.
%There is no 3D dataset that provides \emph{large-scale}, \emph{fine-grained}, \emph{hierarchical} and \emph{instance-level} part annotations, 
%can be addressed by methods leveraging 3D part data.

In this paper, we introduce \PARTS{}: a \emph{consistent, large-scale} dataset on top of ShapeNet~\cite{chang2015shapenet} with \emph{fine-grained, hierarchical, instance-level} 3D part information.
Collecting such fine-grained and hierarchical segmentation is challenging. 
The boundary between fine-grained part concepts are more obscure than defining coarse parts. 
Thus, we define a common set of part concepts by carefully examining the 3D objects to annotate, balancing over several criteria: well-defined, consistent, compact, hierarchical, atomic and complete.
Shape segmentation involves multiple levels of granularity. Coarse parts describe more global semantics and fine-grained parts convey richer geometric and semantic details.
We organize expert-defined part concepts in hierarchical segmentation templates to guide annotation.

PartNet provides a large-scale benchmark for many part-level object understanding tasks. In this paper, we focus on three fundamental and challenging shape segmentation tasks: fine-grained semantic segmentation, hierarchical semantic segmentation, and instance segmentation.
%We create benchmarks for each of these tasks to validate our dataset and to evaluate recent methods for semantic segmentation. 
We benchmark four state-of-the-art algorithms on fine-grained semantic segmentation and propose three baseline methods for hierarchical semantic segmentation.
We propose the task of part instance segmentation using PartNet. By taking advantages of rich shape structures, we propose a novel method that outperforms the existing baseline algorithm by a clear margin.

PartNet contains highly structured, fine-grained and heterogeneous parts.
Our experiments reveals that existing algorithms developed for coarse and homogeneous part understanding cannot work well on PartNet.
First, small and fine-grained parts, \textit{e.g.} door handles and keyboard buttons, are abundant and present new challenges for part recognition.
Second, many geometrically similar but semantically different parts requires more global shape context to distinguish.
Third, understanding the heterogeneous variation of shapes and parts necessitate hierarchical understanding.
We expect that PartNet could serve as a better platform for part-level object understanding in the next few years.

In summary, we make the following contributions:
\begin{itemize}
\vspace{-4pt}
\item 
%We introduce the PartNet dataset, consisting of 573,585 fine-grained part annotations for 26,671 shapes across 24 object categories. To the best of our knowledge, it is the first \textit{large-scale} dataset with \textit{fine-grained, hierarchical, instance-level} part annotations.
We introduce PartNet, consisting of 573,585 fine-grained part annotations for 26,671 shapes across 24 object categories. To the best of our knowledge, it is the first \textit{large-scale} dataset with \textit{fine-grained, hierarchical, instance-level} part annotations;
\vspace{-4pt}
\item We propose three part-level object understanding tasks to demonstrate the usefulness of this data: fine-grained semantic segmentation, hierarchical semantic segmentation, and instance segmentation.
\vspace{-4pt}
\item We benchmark four state-of-the-art algorithms for semantic segmentation and three baseline methods for hierarchical segmentation using PartNet;
\vspace{-4pt}
\item We propose the task of part instance segmentation on PartNet and describe a novel method that outperforms the existing baseline method by a large margin.
\end{itemize}

\begin{figure*}[t]
\centering
   \includegraphics[width=\linewidth]{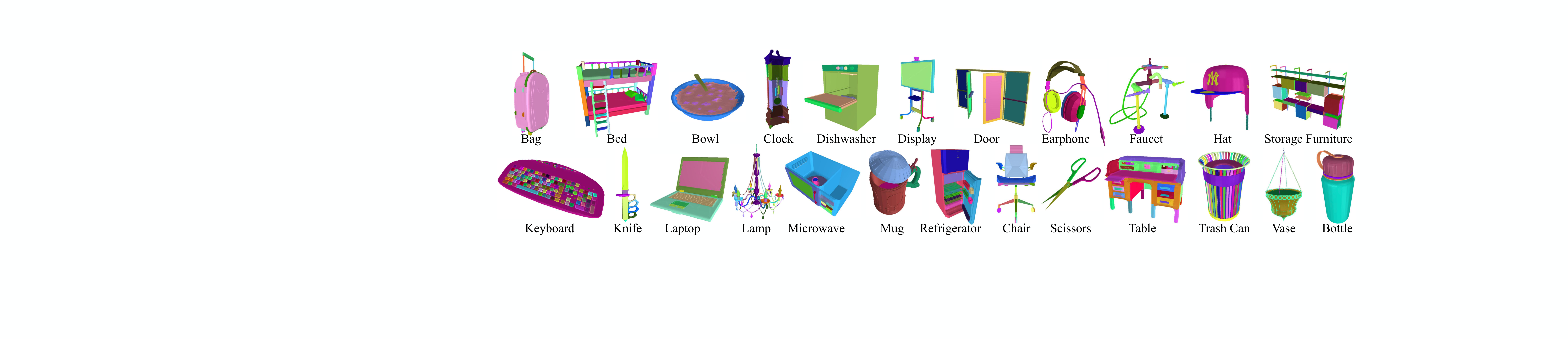}
   \caption{\textbf{PartNet dataset.} We visualize example shapes with fine-grained part annotations for the 24 object categories in PartNet.}
   \label{fig:cat_24}
   \vspace{-1mm}
\end{figure*}

\begin{table*}[t]
    \footnotesize
    \centering
   \begin{tabular}[width=\linewidth]{@{\hspace{0.0cm}}p{0.4cm}|@{\hspace{0.05cm}}p{0.5cm}|@{\hspace{0.05cm}}p{0.15cm}p{0.15cm}p{0.2cm}p{0.25cm} p{0.35cm}p{0.35cm}p{0.25cm}p{0.25cm} p{0.25cm}p{0.15cm}p{0.25cm}p{0.15cm} p{0.15cm}p{0.35cm}p{0.35cm}p{0.1cm} p{0.35cm}p{0.25cm}p{0.25cm}p{0.15cm} p{0.3cm}p{0.3cm}p{0.3cm}p{0.35cm}}
    \Xhline{2\arrayrulewidth}
     & \textbf{All} & \textbf{Bag} & \textbf{Bed} & \textbf{Bott} & \textbf{Bowl} & \textbf{Chair} & \textbf{Clock} & \textbf{Dish} & \textbf{Disp} & \textbf{Door} & \textbf{Ear} & \textbf{Fauc} & \textbf{Hat} & \textbf{Key} & \textbf{Knife} & \textbf{Lamp} & \textbf{Lap} & \textbf{Micro} & \textbf{Mug} & \textbf{Frid} & \textbf{Scis} & \textbf{Stora} & \textbf{Table} & \textbf{Trash} & \textbf{Vase} \\ \Xhline{2\arrayrulewidth}
     \textbf{\#A} & 32537  &  186 & 248 & 519 & 247 & 8176 & 624 & 241 & 1005 & 285 & 285 & 840 & 287 & 210 & 514 & 3408 & 485 & 268 & 252 & 247 & 127 & 2639 & 9906 & 378 & 1160 \\
     \textbf{\#S} & 26671  &  146 & 212 & 464 & 208 & 6400 & 579 & 201 & 954 & 245 & 247 & 708 & 250 & 174 & 384 & 2271 & 453 & 212 & 212 & 207 & 88 & 2303 & 8309 & 340 & 1104 \\
     \textbf{\#M} & 771  &  20 & 18 & 28 & 20 & 77 & 25 & 20 & 26 & 20 & 19 & 60 & 19 & 18 & 57 & 64 & 20 & 28 & 20 & 20 & 20 & 34 & 91 & 19 & 28 \\
     \hline
     \textbf{\#PS} & 480  &  4 & 24 & 12 & 4 & 57 & 23 & 12 & 8 & 8 & 15 & 18 & 8 & 3 & 16 & 83 & 8 & 12 & 4 & 13 & 5 & 36 & 82 & 15 & 10 \\
     \textbf{\#PI} & 573K  &  664 & 9K & 2K & 615 & 176K & 4K & 2K & 7K & 2K & 3K & 8K & 1K & 20K & 3K & 50K & 3K & 2K & 839 & 2K & 981 & 77K & 177K & 8K & 5K \\
     \hline
     \textbf{P\textsubscript{med}} & 14  &  4 & 33 & 5 & 2 & 19 & 5 & 9 & 8 & 7 & 12 & 9 & 4 & 106 & 7 & 12 & 8 & 7 & 3 & 9 & 8 & 24 & 15 & 9 & 4 \\ 
     \textbf{P\textsubscript{max}} & 230  &  7 & 169 & 7 & 4 & 153 & 32 & 16 & 12 & 20 & 14 & 34 & 5 & 127 & 10 & 230 & 8 & 17 & 6 & 33 & 9 & 220 & 214 & 143 & 200 \\ \hline
    \textbf{D\textsubscript{med}} & 3  &  1 & 5 & 2 & 1 & 3 & 3 & 3 & 3 & 3 & 3 & 3 & 2 & 1 & 3 & 5 & 2 & 3 & 1 & 3 & 2 & 4 & 4 & 2 & 2\\
     \textbf{D\textsubscript{max}} & 7  &  1 & 5 & 2 & 1 & 5 & 4 & 3 & 3 & 3 & 3 & 3 & 2 & 1 & 3 & 7 & 2 & 3 & 1 & 3 & 2 & 5 & 6 & 2 & 3 \\ 
     %\hline
    % \textbf{\#L} & 50 & 1 &3 &2 &1 &3 &2 &3 &2  &3 &2 &2 &1 &1 &2 &3 &1  &3 &1 &3 &1 &3 &3 &2 &2 \\ 
    \Xhline{2\arrayrulewidth}
    \end{tabular}
    \vspace{+1mm}
    \caption{\textbf{PartNet statistics.} Row \textbf{\#A}, \textbf{\#S}, \textbf{\#M} respectively show the number of shape annotations, the number of distinct shape instances and the number of shapes that we collect multiple annotations. Row \textbf{\#PS} and \textbf{\#PI} show the number of different part semantics and part instances that we finally collect. Row \textbf{P\textsubscript{med}} and \textbf{P\textsubscript{max}} respectively indicate the median and maximum number of part instances per shape. Row \textbf{D\textsubscript{med}} and \textbf{D\textsubscript{max}} respectively indicate the median and maximum hierarchy depth per shape, with root node as depth 0. 
    %Row \textbf{\#L} shows the number of segmentation levels we define based on the expert defined shape hierarchies for the segmentation benchmarks in Sec \ref{sec:fine_grained_sem_seg} and Sec \ref{sec:ins_seg}.
    }
    \label{tab:partnet_stats}
    \vspace{-3mm}
\end{table*}

%% file: tex/related.tex
Understanding shape parts is a long-standing problem in computer vision and graphics. 
Lacking large-scale annotated datasets, early research efforts evaluated algorithm results qualitatively and conducted quantitative comparison on small sets of 3D models. 
Attene~\etal~\cite{attene2006mesh} compared 5 mesh segmentation algorithms using 11 3D surface meshes and presented side-by-side qualitative comparison. Chen~\etal~\cite{chen2009benchmark} collected 380 surface meshes from 19 object categories with instance-level part decomposition for each shape and proposed quantitative evaluation metrics for shape segmentation. 
Concurrently, Benhabiles~\etal~\cite{benhabiles2009framework} proposed similar evaluation criteria and methodology. 
Kalogerakis~\etal~\cite{kalogerakis2010learning} further assigned semantic labels to the segmented components. 
Shape co-segmentation benchmarks~\cite{wang2012active, hu2012co} were proposed to study co-segmentation among similar shapes.

Recent advances in deep learning have demonstrated the power and efficiency of data-driven methods on 3D shape understanding tasks such as classification, segmentation and generation. 
ShapeNet~\cite{chang2015shapenet} collected a large-scale synthetic 3D CAD models from online open-sourced 3D repositories, including more than 3,000,000 models and 3,135 object categories. 
Yi~\etal~\cite{yi2016scalable} took an active learning approach to annotate the ShapeNet models with semantic segmentation for 31,963 shapes covering 16 object categories. 
In their dataset, each object is usually decomposed into 2$\sim$5 coarse semantic parts. 
%Many recent works \cite{qi2017pointnet, qi2017pointnet++, yi2017syncspeccnn, klokov2017escape, wang2017cnn, li2018so, hermosilla2018monte, wang2018dynamic, wang2018voxsegnet, xu2018spidercnn, su2018splatnet, graham20183d, li2018pointcnn, le2018pointgrid} demonstrated progress on shape semantic segmentation on this dataset. 
PartNet provides more fine-grained part annotations that contains 18 parts per shape on average.

Many recent works studied fine-grained and hierarchical shape segmentation. Yi~\etal~\cite{yi2017learning} leveraged the noisy part decomposition inputs in the CAD model designs and trained per-category models to learn consistent shape hierarchy. Chang~\etal~\cite{chang2018linking} collected 27,477 part instances from 2,278 models covering 90 object categories and studied the part properties related to language. Wang~\etal~\cite{wang2018learning} proposed multi-component labeling benchmark containing 1,016 3D models from ShapeNet~\cite{chang2015shapenet} from 10 object categories with manually annotated fine-grained level part semantics and studied to learn neural networks for grouping and labeling fine-grained part components. PartNet proposes a large-scale dataset with 573,585 fine-grained and hierarchical shape part annotations covering 26,671 models from 24 object categories.

There are also many previous works that attempted to understand parts by their functionality and articulation. Hu~\etal~\cite{hu2016learning} constructed a dataset of 608 objects from 15 object categories annotated with the object functionality and introduced a co-analysis method to learns category-wise object functionality. Hu~\etal~\cite{hu2017learning} proposed a dataset of 368 mobility units with diverse types of articulation and learned to predict part mobility information from a single static segmented 3D mesh. In PartNet, we assign consistent semantic labels that entail such functionality and articulation information for part components within each object category, which potentially makes PartNet support such research.

%% file: tex/framework.tex
%In this section, we describe our data annotation pipeline. 
% We provide expert-defined hierarchical part templates to guarantee labeling consistency among multiple annotators. 
% A simple question-answering 3D GUI is designed to guide the hierarchical annotation.
% We further analyze the human agreement to refine the template details.
% We hire 66 annotators and train them for the annotation task. At least one pass of verification is performed for each annotation. The average annotation time is 8 minutes per shape.
The data annotation is performed in a hierarchical manner. 
Expert-defined hierarchical part templates are provided to guarantee labeling consistency among multiple annotators.
We design a single-thread question-answering 3D GUI to guide the annotation.
%And we also analyze the human agreement to further refine the templates.
We hire 66 professional annotators and train them for the annotation.
The average annotation time per shape is 8 minutes, and at least one pass of verification is performed for each annotation to ensure accuracy.

\subsection{Expert-Defined Part Hierarchy}
\label{sec:part-templates}

\begin{figure*}[t]
    \centering
    \includegraphics[width=\linewidth]{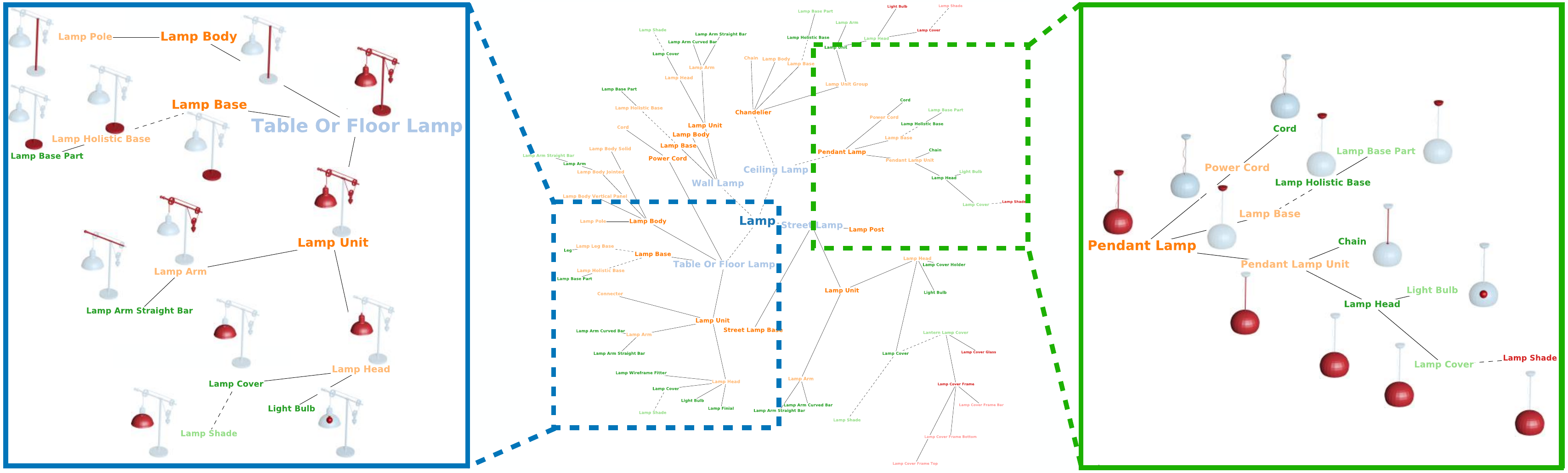}
    \caption{We show the expert-defined hierarchical template for lamp (middle) and the instantiations for a table lamp (left) and a ceiling lamp (right). The And-nodes are drawn in solid lines and Or-nodes in dash lines. The template is deep and comprehensive to cover structurally different types of lamps. In the meantime, the same part concepts, such as light bulb and lamp shade, are shared across the different types.}
    \label{fig:template}
    \vspace{-3mm}
\end{figure*}

Shape segmentation naturally involves hierarchical understanding. 
People understand shapes at different segmentation granularity. 
Coarse parts convey global semantics while fine-grained parts provide more detailed understanding. 
Moreover, fine-grained part concepts are more obscure to define than coarse parts.
Different annotators have different knowledge and background so that they may name parts differently when using free-form annotation~\cite{chang2018linking}.
To address the issues, we introduce And-Or-Graph-style hierarchical templates and collect part annotations according to the pre-defined templates.

% Though seemingly simple, designing  hierarchical part templates to cover all variations of shapes and parts in a category is a non-trivial task. 
% There lacks well-acknowledged rules of thumb to define good templates. 
Due to the lack of well-acknowledged rules of thumb to define good templates, the task of designing hierarchical part templates for a category becomes a non-trivial task. 
Furthermore, the requirement for the designed template to cover all variations of shapes and parts, makes the problem more challenging. 
Below we summarize the criteria that we use to guide our template design:
\begin{itemize}
    \vspace{-4pt}
   % \item \textbf{Definitiveness:} Part concepts are well-defined and well distinguished from similar parts; 
    \item \textbf{Well-defined}: Part concepts are well-delineated such that parts are identifiable by multiple annotators;
   % Definitive?
    \vspace{-8pt}
    \item \textbf{Consistent:} Part concepts are shared and reused across different parts, shapes and object categories; 
    \vspace{-8pt}
    \item \textbf{Compact:} There is no unnecessary part concept and part concepts are reused when it is possible;
    \vspace{-8pt}
    \item \textbf{Hierarchical:} Part concepts are organized in a taxonomy to cover both coarse and fine-grained parts;
    \vspace{-8pt}
    \item \textbf{Atomic:} Leaf nodes in the part taxonomy consist of primitive, non-decomposable shapes;
    \vspace{-8pt}
    \item \textbf{Complete:} The part taxonomy covers a heterogeneous variety of shapes as completely as possible.
\end{itemize}\vspace{-4pt}
Guided by these general principles, we build an And-Or-Graph-style part template for each object category. The templates are defined by experts after examining a broad variety of objects in the category. Each template is designed in a hierarchical manner from the coarse semantic parts to the fine-grained primitive-level components. Figure~\ref{fig:template} (middle) shows the lamp template. And-nodes segment a part into small subcomponents. Or-nodes indicate subcategorization for the current part. The combination of And-nodes and Or-nodes allows us to cover structurally different shapes using the same template while sharing as much common part labels as possible. As in Figure~\ref{fig:template} (left) and (right), both table lamps and ceiling lamps are explained by the same template through the first-level Or-node for lamp types.

Despite the depth and comprehensiveness of these templates, it is still impossible to cover all cases. Thus, we allow our annotators to improve upon the structure of the template and to annotate parts that are out of the scope of our definition. We also conduct template refinements to resolve part ambiguity after we obtain the data annotation according to the original templates. To systematically identify ambiguities, we reserve a subset of shapes from each class and collect multiple human annotations for each shape. We compute the confusion matrix among different annotators and address data inconsistencies. For example, we merge two concepts with high confusion scores or remove a part if it is frequently segmented in the wrong way. We provide more details about this in the supplementary material.

\subsection{Annotation Interface}
\label{sec:annotation-interface}

%\begin{figure}[t!]
%    \centering
%    \includegraphics[width=\linewidth]{fig/annotation/interface4}
%    \caption{\textbf{3D Annotation Interface.} 
%    % We present the 3D models in along with the question-answering workflow (middle section) which guides the annotator to obtain the hierarchical segmentation (left section) step by step. We show the grey 3D mesh and textured 2D snapshots on the right section. At the bottom of the question-answering section, we provide detailed text description and visual examples to explain the part definition.
%    The interface shows an unlabeled 3D model in grey 3D mesh (right top) \zhe{do you want to include a figure with grey mesh?} along with the question-answering workflow (center top), which guides the annotator in a hierarchical segmentation (left) step by step. Detailed text descriptions and visual examples are shown at the center bottom, to explain the part definition, and textured 2D snapshots of the model are displayed at the right bottom.
%    }
%    \label{fig:interface}
%\end{figure}

\begin{figure*}[t]
    \centering
    \includegraphics[width=\linewidth]{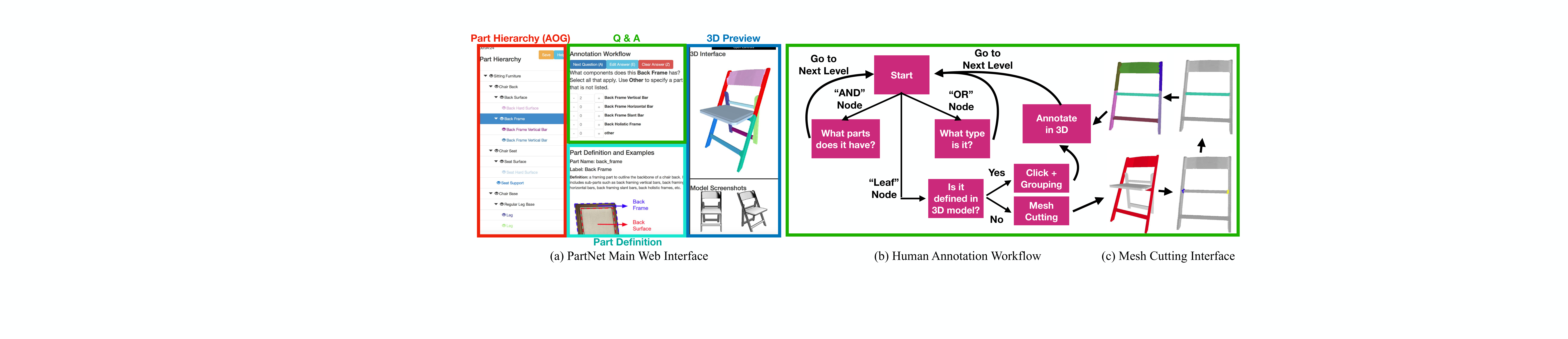}
    \vspace{-6mm}
    \caption{We show our annotation interface with its components, the proposed question-answering workflow and the mesh cutting interface.}
    \label{fig:anno_workflow}
    \vspace{-3mm}
\end{figure*}

% We expose the templates to the annotators by converting the templates to a single-thread question-answering workflow. Figure~\ref{fig:interface} shows our annotation interface. 
% At each stage, the annotator indicates the number of sub-parts for that node.  
% When the leaf node is reached, the annotator then uses the 3D viewer to click and group the subparts of the shape corresponding to the leaf node.  To help the annotators understand the part specification, we provided visual examples of each of the parts. In addition, we also prepared comprehensive video tutorials.
Figure~\ref{fig:anno_workflow} (a) shows our web-based annotation interface. Based on the template hierarchy, the annotation process is designed to be a single-thread question-answering workflow, traversing the template graph in a depth-first manner, as shown in Figure~\ref{fig:anno_workflow} (b). Starting from the root node, the annotator is asked a sequence of questions. The answers automatically construct the final hierarchical segmentation for the current shape instance.
For each question, the annotator is asked to mark the number of subparts (And-node) or pick one among all subtypes (Or-node) for a given part.  
For each leaf node part, the annotator annotates the part geometry in the 3D interface.
To help them understand the part definition and specification, we provide rich textual definitions and visual examples for each part. In addition, our interface supports cross-section and visibility control to annotate the interior structure of a 3D model.

% 3D CAD models often come with the original mesh sub-groups and part information. Many of them are sufficiently detailed enough to determine the final segmentation as we need, but they exhibit significant variation in depth and inconsistencies between shapes. To speed up the annotation, we use the existing part structures from the CAD model designs as much as possible. We present the annotators with the most fine-grained segmentation that come with the original CAD models and allow the users to simply click multiple pieces and label them as one part for the final segmentation. 
% However, the original part segments may be too coarse for the final segmentation we want. We allow the user to perform mesh cutting to split large parts into smaller fine-grained parts following \cite{chen2009benchmark}. As shown in Figure~\ref{fig:mesh_cutting}, the mesh cutting is achieved by allowing the annotators to draw several lines at the part boundary to split up the mesh into multiple sub-components. This requires the part mesh to be water-tight \textcolor{blue}{a watertight} manifold.   We use \cite{huang2018robust} to generate manifold part for ShapeNet models.

The collected 3D CAD models often include original mesh subgroups and part information.
Some of the grouping information is detailed enough to determine the final segmentation we need.
Considering this, we provide the annotators with the original groupings at the beginning of the annotation, to speed up annotation.
The annotators can simply select multiple predefined pieces to form a part of the final segmentation. 
We also provide mesh cutting tools to split large pieces into smaller ones following~\cite{chen2009benchmark}, when the original groupings are coarser than the desired segmentation, as shown in Figure~\ref{fig:anno_workflow} (c).
The annotators draw boundary lines on the remeshed watertight surface~\cite{huang2018robust} and the mesh cutting algorithm automatically splits the mesh into multiple smaller subcomponents. 

%\begin{figure}[t!]
%    \centering
%    \includegraphics[width=\linewidth]{fig/annotation/mesh_cutting}
%    \caption{\textbf{Mesh Cutting Interface.} To cut a part into smaller sub-components, the annotator first selects the part (a), and then draw several lines by adding pivot points on the remeshed part manifold to form boundaries between different parts (b). Then, the system splits the mesh into two parts: one in grey and one in blue (c). By iterating the above steps, we obtain the final segmentation (d).}
%    \label{fig:mesh_cutting}
%\end{figure}

In contrast to prior work, our UI is designed for operating directly on 3D models and collecting fine-grained and hierarchical part instances. Compared to Yi~\etal~\cite{yi2016scalable} where the annotation is performed in 2D, our approach allows the annotators to directly annotate on the 3D shapes and thus be able to pick up more subtle part details that are hidden from 2D renderings. Chang~\etal~\cite{chang2018linking} proposes a 3D UI that paints regions on mesh surfaces for part labeling.  However, their interface is limited to existing over-segmentations on part components and does not support hierarchical annotations.
%We take advantage of the pre-existing part components in the original 3D CAD designs to speed up the annotation speed.
%The proposed template hierarchy and the coupled user interface enables annotating directly on the 3D shape, compared to the work by Yi \etal~\cite{yi2016scalable}, and grouping or splitting mesh on the pre-defined segments, compared to the work by Chang et al \cite{chang2018linking}.
%Overall, the advantages of the annotation system lead to more consistent and precise 3D part annotations.  

%% file: tex/data.tex
The final PartNet dataset provides fine-grained and hierarchical instance-level part segmentation annotation for $26,671$ shapes with $573,585$ part instances from $24$ object categories. Most of the shapes and object categories are from ShapeNetCore~\cite{chang2015shapenet}. We supplement $3$ object categories that are commonly present in indoor scenes (\textit{i.e.} scissors, refrigerators, and doors) and augment $7$ of the existing categories with more 3D models from 3D Warehouse\footnote{\href{https://3dwarehouse.sketchup.com}{https://3dwarehouse.sketchup.com}}. 
%Most object categories have at least $100$ shape instances to support data-hungry algorithms, while for scissors we only have $88$.

Figure~\ref{fig:cat_24} and Table~\ref{tab:partnet_stats} show the PartNet data and statistics. More visualization and statistics are included in supplemental material. Our templates define hierarchical segmentation with $3$ depth in median and $7$ maximum. In total, we annotate $573,585$ fine-grained part instances, with a median of $14$ parts per shape and a maximum of $230$. To study annotation consistency, we also collect a subset of $771$ shapes and ask for multiple annotations per shape.

%% file: tex/exp.tex
We benchmark three part-level object understanding tasks using PartNet: fine-grained semantic segmentation, hierarchical semantic segmentation and instance segmentation. 
Four state-of-the-art algorithms for semantic segmentation are evaluated and three baseline methods are proposed for hierarchical segmentation.
Moreover, we propose a novel method for instance segmentation that outperforms the existing baseline method.

% \paragraph{Data Preparation. }
\vspace{-0.3em}
{\flushleft \bf Data Preparation.}
% In this paper, we learn part semantics that can be fully determined by the shape geometry\footnote{Although 3D models in ShapeNet \cite{chang2015shapenet} come with face normal, textures, material and other information, there is no guarantee for the quality of such information. Thus, we leave this as a future work.}. We remove the parts from evaluation that require additional information to detect. For example, the glass parts on the cabinet doors require opacity to tell and the buttons on microwaves may be indistinguishable from the main frame if no color information is provided. We also remove the rarely annotated parts from the evaluation since there are not enough shapes containing the parts for both training and testing. For instance, we have only $11$ lava lamps among $3,408$ lamps, which is too few for the benchmarks. We summarize the final sets of parts we use in the supplemental material.
%Although 3D models in ShapeNet~\cite{chang2015shapenet} come with additional information, \eg, face normal, textures and material information, there is no guarantee of their quality.
In this section, we only consider parts that can be fully determined by their shape geometry\footnote{Although 3D models in ShapeNet \cite{chang2015shapenet} come with face normal, textures, material and other information, there is no guarantee for the quality of such information. Thus, we leave this as a future work.}. 
% And we leave the task using additional information as the future work. (can be moved to future work section)
We ignore the parts in evaluation that require additional information to identify, such as  
the glass parts on the cabinet doors which opacity is needed to identify, and the buttons on microwaves that texture information is desired to distinguish it from the main frame. 
We also remove rarely appeared parts from the evaluation, as the lacking of samples is insufficient for training and evaluating networks. 
%For instance, there are only $11$ lava lamps \zhe{is it a part or a class?} among $3,408$ lamps. 
%We summarize more details in the supplementary material.

% We sample $10,000$ point clouds from the CAD models using furthest point sampling and use the 3D locations as the neural network inputs. We split the entire dataset into train, validation and test splits with ratio $70\%, 10\%, 20\%$ for each object category. We leave out the shapes with multiple human annotations in both training and testing. They are used in the template refinement stage and potentially a testing source for human performance for the tasks.
We sample $10,000$ points from each CAD model with furthest point sampling and use the 3D coordinates as the neural network inputs for all the experiments in the paper. 
The proposed dataset is split into train, validation and test sets with the ratio 70\%: 10\%: 20\%. 
The shapes with multiple human annotations are not used in the experiments.

\subsection{Fine-grained Semantic Segmentation}
\label{sec:fine_grained_sem_seg}

\begin{table*}[th!]
    \footnotesize
    \centering
   \begin{tabular}[width=\linewidth]{@{\hspace{0.1cm}}p{0.3cm}|@{\hspace{0.1cm}}p{0.35cm}|@{\hspace{0.1cm}}p{0.15cm}p{0.15cm}p{0.2cm}p{0.25cm} p{0.35cm}p{0.35cm}p{0.25cm}p{0.25cm} p{0.25cm}p{0.15cm}p{0.25cm}p{0.15cm} p{0.15cm}p{0.35cm}p{0.35cm}p{0.15cm} p{0.35cm}p{0.25cm}p{0.25cm}p{0.15cm} p{0.35cm}p{0.35cm}p{0.35cm}p{0.35cm}}
    \Xhline{2\arrayrulewidth}
     & \textbf{Avg} & \textbf{Bag} & \textbf{Bed} & \textbf{Bott} & \textbf{Bowl} & \textbf{Chair} & \textbf{Clock} & \textbf{Dish} & \textbf{Disp} & \textbf{Door} & \textbf{Ear} & \textbf{Fauc} & \textbf{Hat} & \textbf{Key} & \textbf{Knife} & \textbf{Lamp} & \textbf{Lap} & \textbf{Micro} & \textbf{Mug} & \textbf{Frid} & \textbf{Scis} & \textbf{Stora} & \textbf{Table} & \textbf{Trash} & \textbf{Vase} \\ \Xhline{2\arrayrulewidth}
     \textbf{P1} & 57.9 & 42.5 & 32.0 & 33.8 & 58.0 & 64.6 & 33.2 & 76.0 & 86.8 & \textbf{64.4} & 53.2 & 58.6 & 55.9 & 65.6 & 62.2 & 29.7 & 96.5 & 49.4 & 80.0 & 49.6 & 86.4 & 51.9 & 50.5 & 55.2 & 54.7 \\
     \textbf{P2} & 37.3 & $\;\;-$ & 20.1 & $\;\;-$ & $\;\;-$ & 38.2 & $\;\;-$ & 55.6 & $\;\;-$ & 38.3 & $\;\;-$ & $\;\;-$ & $\;\;-$ & $\;\;-$ & $\;\;-$ & 27.0 & $\;\;-$ & 41.7 & $\;\;-$ & 35.5 & $\;\;-$ & 44.6 & 34.3 & $\;\;-$ & $\;\;-$ \\
     \textbf{P3} & 35.6 & $\;\;-$ & 13.4 & 29.5 & $\;\;-$ & 27.8 & 28.4 & 48.9 & 76.5 & 30.4 & 33.4 & 47.6 & $\;\;-$ & $\;\;-$ & 32.9 & 18.9 & $\;\;-$ & 37.2 & $\;\;-$ & 33.5 &  $\;\;-$ & 38.0 & 29.0 & 34.8 & 44.4 \\ \hline
     \textbf{Avg} & 51.2 & 42.5 & 21.8 & 31.7 & 58.0 & 43.5 & 30.8 & 60.2 & 81.7 & 44.4 & 43.3 & 53.1 & 55.9 & 65.6 & 47.6 & 25.2 & 96.5 & 42.8 & 80.0 & 39.5 & 86.4 & 44.8 & 37.9 & 45.0 & 49.6 \\ \Xhline{2\arrayrulewidth}
     \textbf{P\textsuperscript{+}1} & \textbf{65.5} & 59.7 & 51.8 & 53.2 & 67.3 & 68.0 & \textbf{48.0} & 80.6 & 89.7 & 59.3 & \textbf{68.5} & 64.7 & 62.4 & 62.2 & \textbf{64.9} & \textbf{39.0} & 96.6 & 55.7 & 83.9 & 51.8 & 87.4 & 58.0 & \textbf{69.5} & 64.3 & \textbf{64.4} \\
     \textbf{P\textsuperscript{+}2} & 44.5 & $\;\;-$ & 38.8 & $\;\;-$ & $\;\;-$ & 43.6 & $\;\;-$ & 55.3 & $\;\;-$ & 49.3 & $\;\;-$ & $\;\;-$ & $\;\;-$  & $\;\;-$ & $\;\;-$ &  \textbf{32.6} & $\;\;-$ & 48.2 & $\;\;-$ & 41.9 & $\;\;-$ & 49.6 & \textbf{41.1} & $\;\;-$ & $\;\;-$ \\
     \textbf{P\textsuperscript{+}3} & 42.5 & $\;\;-$ & 30.3 & 41.4 & $\;\;-$ & 39.2 & \textbf{41.6} & 50.1 & 80.7 & 32.6 & 38.4 & 52.4 & $\;\;-$ & $\;\;-$ & \textbf{34.1} & \textbf{25.3} & $\;\;-$ & 48.5 & $\;\;-$ & 36.4 & $\;\;-$ & 40.5 & \textbf{33.9} & 46.7 & 49.8 \\ \hline
     \textbf{Avg} & 58.1 & 59.7 & 40.3 & \textbf{47.3} & 67.3 & 50.3 & \textbf{44.8} & 62.0 & 85.2 & 47.1 & 53.5 & 58.6 & 62.4 & 62.2 & \textbf{49.5} & \textbf{32.3} & 96.6 & 50.8 & 83.9 & 43.4 & 87.4 & 49.4 & \textbf{48.2} & 55.5 & 57.1 \\ \Xhline{2\arrayrulewidth}
     \textbf{S1} & 60.4 & 57.2 & 55.5 & \textbf{54.5} & \textbf{70.6} & 67.4 & 33.3 & 70.4 & 90.6 & 52.6 & 46.2 & 59.8 & 63.9 & 64.9 & 37.6 & 30.2 & \textbf{97.0} & 49.2 & 83.6 & 50.4 & 75.6 & \textbf{61.9} & 50.0 & 62.9 & 63.8 \\
     \textbf{S2} & 41.7 & $\;\;-$ & 40.8 & $\;\;-$ & $\;\;-$ & 39.6 & $\;\;-$ & 59.0 & $\;\;-$ & 48.1 & $\;\;-$ & $\;\;-$ & $\;\;-$ & $\;\;-$ & $\;\;-$ & 24.9 & $\;\;-$ & 47.6 & $\;\;-$ & 34.8 & $\;\;-$ & 46.0 & 34.5 & $\;\;-$ & $\;\;-$ \\
     \textbf{S3} & 37.0 & $\;\;-$ & 36.2 & 32.2 & $\;\;-$ & 30.0 & 24.8 & 50.0 & 80.1 & 30.5 & 37.2 & 44.1 & $\;\;-$ & $\;\;-$ & 22.2 & 19.6 & $\;\;-$ & 43.9 & $\;\;-$ & 39.1 & $\;\;-$ & 44.6 & 20.1 & 42.4 & 32.4 \\ \hline
     \textbf{Avg} & 53.6 & 57.2 & 44.2 & 43.4 & \textbf{70.6} & 45.7 & 29.1 & 59.8 & 85.4 & 43.7 & 41.7 & 52.0 & 63.9 & 64.9 & 29.9 & 24.9 & \textbf{97.0} & 46.9 & 83.6 & 41.4 & 75.6 & 50.8 & 34.9 & 52.7 & 48.1 \\ \Xhline{2\arrayrulewidth}
     \textbf{C1} & 64.3 & \textbf{66.5} & \textbf{55.8} & 49.7 & 61.7 & \textbf{69.6} & 42.7 & \textbf{82.4} & \textbf{92.2} & 63.3 & 64.1 & \textbf{68.7} & \textbf{72.3} & \textbf{70.6} & 62.6 & 21.3 & \textbf{97.0} & \textbf{58.7} & \textbf{86.5} & \textbf{55.2} & \textbf{92.4} & 61.4 & 17.3 & \textbf{66.8} & 63.4 \\
     \textbf{C2} & \textbf{46.5} & $\;\;-$ & \textbf{42.6} & $\;\;-$ & $\;\;-$ & \textbf{47.4} & $\;\;-$ & \textbf{65.1} & $\;\;-$ & \textbf{49.4} & $\;\;-$ & $\;\;-$ & $\;\;-$ & $\;\;-$ & $\;\;-$ & 22.9 & $\;\;-$ & \textbf{62.2} & $\;\;-$ & \textbf{42.6} & $\;\;-$ &\textbf{57.2} & 29.1 & $\;\;-$ & $\;\;-$ \\
     \textbf{C3} & \textbf{46.4} & $\;\;-$ & \textbf{41.9} & \textbf{41.8} & $\;\;-$ & \textbf{43.9} & 36.3 & \textbf{58.7} & \textbf{82.5} & \textbf{37.8} & \textbf{48.9} & \textbf{60.5} & $\;\;-$ & $\;\;-$ & \textbf{34.1} & 20.1 & $\;\;-$ & \textbf{58.2} & $\;\;-$ & \textbf{42.9} & $\;\;-$ & \textbf{49.4} & 21.3 & \textbf{53.1} & \textbf{58.9} \\ \hline
     \textbf{Avg} & \textbf{59.8} & \textbf{66.5} & \textbf{46.8} & 45.8 & 61.7 & \textbf{53.6} & 39.5 & \textbf{68.7} & \textbf{87.4} & \textbf{50.2} & \textbf{56.5} & \textbf{64.6} & \textbf{72.3} & \textbf{70.6} & 48.4 & 21.4 & \textbf{97.0} & \textbf{59.7} & \textbf{86.5} & \textbf{46.9} & \textbf{92.4} & \textbf{56.0} & 22.6 & \textbf{60.0} & \textbf{61.2} \\ \Xhline{2\arrayrulewidth}
    \end{tabular}
    \vspace{-1mm}
    \caption{\textbf{Fine-grained semantic segmentation results (part-category mIoU \%).} Algorithm \textbf{P}, \textbf{P\textsuperscript{+}}, \textbf{S} and \textbf{C} refer to PointNet \cite{qi2017pointnet}, PointNet++ \cite{qi2017pointnet++}, SpiderCNN \cite{xu2018spidercnn} and PointCNN \cite{li2018pointcnn}, respectively. The number \textbf{1}, \textbf{2} and \textbf{3} refer to the three levels of segmentation: coarse-, middle- and fine-grained. We put short lines for the levels that are not defined.}
    \label{tab:sem_seg_per_part_category_result}
    \vspace{-3mm}
\end{table*}

% Many previous works \cite{qi2017pointnet, qi2017pointnet++, yi2017syncspeccnn, klokov2017escape, wang2017cnn, li2018so, hermosilla2018monte, wang2018dynamic, wang2018voxsegnet, xu2018spidercnn, su2018splatnet, graham20183d, li2018pointcnn, le2018pointgrid} are pushing the state-of-the-art performance on coarse-level semantic segmentation on the ShapeNet Part dataset \cite{chang2015shapenet, yi2016scalable}. Based on our proposed PartNet dataset, we propose a more challenging task of 3D shape fine-grained semantic segmentation.
Recent advances of 3D semantic segmentation~\cite{qi2017pointnet, qi2017pointnet++, yi2017syncspeccnn, klokov2017escape, wang2017cnn, li2018so, hermosilla2018monte, wang2018dynamic, wang2018voxsegnet, xu2018spidercnn, su2018splatnet, graham20183d, li2018pointcnn, le2018pointgrid} have accomplished promising achievement in coarse-level segmentation on the ShapeNet Part dataset~\cite{chang2015shapenet, yi2016scalable}.
However, few work focus on the fine-grained 3D semantic segmentation, due to the lack of large-scale fine-grained dataset.
With the help of the proposed PartNet dataset, researchers can now work on this more challenging task with little overhead.

% Fine-grained semantic segmentation requires recognizing and distinguishing small and similar semantic parts. For example, door handles are usually very small (i.e. only $77$ out of $10,000$ points on average in PartNet) but important semantic parts on doors for vision systems to recognize. A bed may have many geometrically similar parts such as side vertical bars, post bars and base legs. To correctly recognize the subtle part details, segmentation systems need to learn strong discriminative features locally and be capable of understanding the part in the context of the whole shape.
Fine-grained 3D semantic segmentation requires recognizing and distinguishing small and similar semantic parts. 
For example, door handles are usually small, $77$ out of $10,000$ points on average in PartNet, but semantically important on doors. 
Beds have several geometrically similar parts such as side vertical bars, post bars and base legs. 
To recognize the subtle part details, segmentation systems need to understand them locally, through discriminative features, and globally, in the context of the whole shape.

%\paragraph{Benchmark Algorithms.} 
%\vspace{-0.5em}
{\flushleft \bf Benchmark Algorithms.}
%\paragraph{Benchmark Algorithms.}
% We benchmark four state-of-the-art semantic segmentation algorithms\footnote{There are many other algorithm candidates: \cite{yi2017syncspeccnn, klokov2017escape, wang2017cnn, li2018so, hermosilla2018monte, wang2018dynamic, wang2018voxsegnet, su2018splatnet, graham20183d, le2018pointgrid}. We will host an online leaderboard to report the performance for these algorithms.} on the fine-grained PartNet segmentation: PointNet \cite{qi2017pointnet}, PointNet++ \cite{qi2017pointnet++}, SpiderCNN \cite{xu2018spidercnn} and PointCNN \cite{li2018pointcnn}. PointNet \cite{qi2017pointnet} proposes a network that takes unordered point sets as inputs and extract features for shape classification and segmentation. The follow-up work PointNet++ \cite{qi2017pointnet++} further describes a hierarchical feature extraction scheme to learn better local geometric features. SpiderCNN \cite{xu2018spidercnn} extends the traditional convolution operations on 2D images to 3D point clouds by parametrizing a family of convolutional kernels. PointCNN \cite{li2018pointcnn} proposes to learn $\mathcal{X}$-transformation to permute the unordered points into latent canonical order and apply $\mathcal{X}$-convolution operations on the canonical points. We refer the readers to their papers for details.
We benchmark four state-of-the-art semantic segmentation algorithms on the fine-grained PartNet segmentation: PointNet~\cite{qi2017pointnet}, PointNet++~\cite{qi2017pointnet++}, SpiderCNN~\cite{xu2018spidercnn} and PointCNN~\cite{li2018pointcnn}\footnote{There are many other algorithm candidates: \cite{yi2017syncspeccnn, klokov2017escape, wang2017cnn, li2018so, hermosilla2018monte, wang2018dynamic, wang2018voxsegnet, su2018splatnet, graham20183d, le2018pointgrid}. We will host an online leadboard to report the performances.}. 
PointNet~\cite{qi2017pointnet} takes unordered point sets as inputs and extracts features for shape classification and segmentation. 
To better learn local geometric features, the follow-up work PointNet++~\cite{qi2017pointnet++} proposes a hierarchical feature extraction scheme. 
SpiderCNN~\cite{xu2018spidercnn} extends traditional convolution operations on 2D images to 3D point clouds by parameterizing a family of convolutional filters. 
To organize the unordered points into latent canonical order, PointCNN~\cite{li2018pointcnn} proposes to learn $\mathcal{X}$-transformation, and applies $\mathcal{X}$-convolution operations on the canonical points.
%We refer the readers to their papers for details.

We train the four methods on the dataset, using the default network architectures and hyperparameters described in their papers. 
Instead of training a single network for all object categories as done in most of these papers, we train a network for each category at each segmentation level. 
We input only the 3D coordinates for fair comparison\footnote{PointNet++~\cite{qi2017pointnet++} and SpiderCNN~\cite{xu2018spidercnn} use point normals as additional inputs. For fair comparison, we only input the 3D coordinates.} and train the networks until convergence. 
More training details are described in the supplementary material.

% We train the four methods following the default network architectures and network hyperparameters in the papers. Instead of jointly training a single across all object categories as done in most of these papers, we train separate networks for different categories and segmentation levels. In the original papers, PointNet++ \cite{qi2017pointnet++} and SpiderCNN \cite{xu2018spidercnn} use point normals as inputs. For fair comparison, we only provide the point locations as inputs to all the methods. We train all the networks until convergence. More training details are in the supplemental material.
%
%

\begin{figure}[t!]
    \centering
    \includegraphics[width=\linewidth]{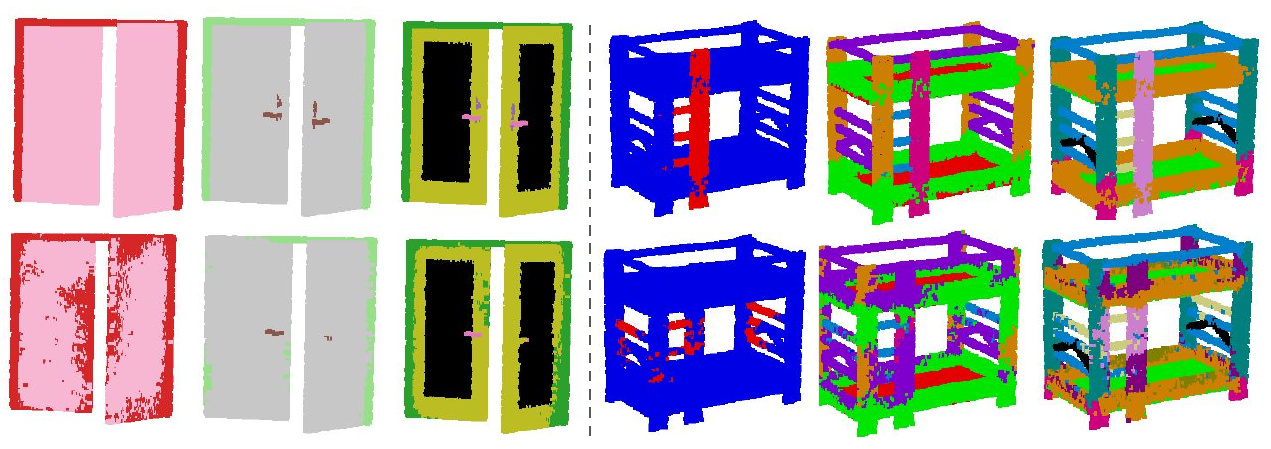}
    \caption{\textbf{Qualitative results for semantic segmentation.} The top row shows the ground-truth and the bottom row shows the PointCNN prediction. The black points indicate unlabeled points.}
    \label{fig:sem_seg_result}
    \vspace{-3mm}
\end{figure}

%\vspace{-0.5em}
{\flushleft \bf Evaluation and Results.}
%\paragraph{Evaluation and Results.}
% To evaluate the performance of the segmentation algorithms at different segmentation granularity, we pick three levels of segmentation for each object category: coarse, middle and fine-grained. The coarse level of segmentation approximately corresponds to the ShapeNet coarse segmentation dataset \cite{yi2016scalable}, while the fine-grained level corresponds to the leaf levels in the segmentation hierarchies. For structurally rich hierarchies, we also select one middle level in between. Among all 24 object categories, all of them have the coarse levels, while 9 have the middle levels and 17 have the fine-grained levels.  We evaluate the performance on all 24 object categories and 50 segmentation levels.
%To understand segmentation at different granularities, 
We evaluate the algorithms at three segmentation levels for each object category: coarse-, middle- and fine-grained. 
The coarse level approximately corresponds to the granularity in Yi~\etal~\cite{yi2016scalable}. The fine-grained level refers to the segmentation down to leaf levels in the segmentation hierarchies. 
For structurally deep hierarchies, we define the middle level in between. 
Among 24 object categories, all of them have the coarse levels, while 9 have the middle levels and 17 have the fine levels. 
Overall, we define 50 segmentation levels for 24 object categories.

In Table \ref{tab:sem_seg_per_part_category_result}, we report the semantic segmentation performances at multiple levels of granularity on PartNet. 
We use the mean Intersection-over-Union (mIoU) scores as the evaluation metric. 
After removing unlabeled ground-truth points, for each object category, we first calculate the IoU between the predicted point set and the ground-truth point set for each semantic part category across all test shapes. 
Then, we average the per-part-category IoUs to compute the mIoU for the object category. 
We further calculate the average mIoU across different levels for each object category and finally report the average cross all object categories.

%We use the mean Intersection-over-Union (mIoU) score as the evaluation metric.
%After removing unlabeled ground-truth points, we calculate the mIoU between the predicted point set and the ground-truth point set for each semantic part category across all test shapes. 
%Then, we compute per-object-category mIoUs by averaging their per-part-category mIoUs. 
%Moreover, we calculate granularity-level mIoU for each object category and the average mIoUs cross all object categories.
%Then, for each object category, we average the per-part-category IoUs to compute the mIoU. We further calculate the average mIoU across different levels for each object category and finally report the average cross all object categories.
%Results of the performance on PartNet are presented in Table~\ref{tab:sem_seg_per_part_category_result}.

% We observe clear performance drops for all the four algorithms when the parts go from the coarse levels to the fine-grained levels. Figure \ref{fig:sem_seg_result} shows qualitative semantic segmentation results for PointCNN \cite{li2018pointcnn}. On the door example, the door handle is very small and thus challenging to detect. On the bed example, we observe clearly that the network fails to distinguish visually similar but semantically different parts, such as the stair steps and the connecting bars on the bed frame. Future works may investigate how to learn more distinctive features that better captures both local geometry and global context to correctly recognize small and similar parts.

Unsurprisingly, performance for all four algorithms drop by a large margin from the coarse level to the fine-grained level. 
Figure~\ref{fig:sem_seg_result} shows qualitative results from PointCNN. 
The method does not perform well on small parts, such as the door handle on the door example, and visually similar parts, such as stair steps and the horizontal bars on the bed frame.
How to learn discriminative features that better capture both local geometry and global context for these issues would be an interest topic for future works.

\subsection{Hierarchical Semantic Segmentation}
\label{sec:hier_sem_seg}

\begin{table*}[th!]
    \footnotesize
    \centering
     \begin{tabular}[width=\linewidth]{@{\hspace{0.15cm}}p{1.4cm}|@{\hspace{0.15cm}}p{0.4cm}|@{\hspace{0.15cm}}p{0.45cm}p{0.45cm}p{0.45cm}p{0.45cm} p{0.45cm}p{0.45cm}p{0.45cm}p{0.45cm} p{0.45cm}p{0.45cm}p{0.45cm}p{0.45cm} p{0.45cm}p{0.45cm}p{0.45cm}p{0.45cm} p{0.5cm}}
    \Xhline{2\arrayrulewidth}
     & \textbf{Avg} & \textbf{Bed} & \textbf{Bott} & \textbf{Chair} & \textbf{Clock} & \textbf{Dish} & \textbf{Disp} & \textbf{Door} & \textbf{Ear} & \textbf{Fauc} & \textbf{Knife} & \textbf{Lamp} & \textbf{Micro} & \textbf{Frid} & \textbf{Stora} & \textbf{Table} & \textbf{Trash} & \textbf{Vase} \\ \Xhline{2\arrayrulewidth}
     \textbf{Bottom-Up} & 51.2 & 40.8 & 56.1 & 47.2 & 38.3 & \textbf{61.5} & 84.1 & \textbf{52.6} & 54.3 & \textbf{63.4} & 52.3 & 36.8 & 48.2 & 41.0 & 46.8 & 38.3 & \textbf{53.6} & 54.4 \\
     \textbf{Top-Down} & 50.8 & 41.1 & \textbf{56.2} & 46.5 & 34.3 & 54.5 & \textbf{84.7} & 50.6 & \textbf{59.5} & 61.4 & \textbf{55.6} & 37.1 & \textbf{48.8} & \textbf{41.6} & 45.2 & 37.0 & 53.5 & \textbf{55.6} \\
     \textbf{Ensemble} & \textbf{51.7} & \textbf{42.0} & 54.7 & \textbf{48.1} & \textbf{44.5} & 58.8 & \textbf{84.7} & 51.4 & 57.2 & 61.9 & 51.9 & \textbf{37.6} & 47.5 & 41.4 & \textbf{47.3} & \textbf{44.0} & 52.8 & 53.1 \\ \Xhline{2\arrayrulewidth}
    \end{tabular}
    \vspace{+1mm}
    \caption{\textbf{Hierarchical segmentation results (part-category mIoU \%).} We present the hierarchical segmentation performances for three baseline methods: bottom-up, top-down and ensemble. We conduct experiments on 17 out of 24 categories with tree depth bigger than 1.}
    \label{tab:hier_seg_per_part_category_result}
    \vspace{-3mm}
\end{table*}

% Shape part segmentation has a hierarchical nature. From coarse semantics to fine-grained details, understanding 3D objects in a hierarchical way develops a holistic and comprehensive reasoning on the shape components. For this task, we study hierarchical semantic segmentation that predicts semantic part labels on the entire expert-defined shape part hierarchies that cover both coarse and fine-grained part concepts using PartNet.
Shape segmentation is hierarchical by its nature. 
From coarse semantics to fine-grained details, hierarchical understanding on 3D objects develops a holistic and comprehensive reasoning on the shape components. 
For this purpose, we study hierarchical semantic segmentation problem that predicts semantic part labels in the entire shape hierarchies that cover both coarse- and fine-grained part concepts.

% The key question to answer towards hierarchical segmentation is how to leverage the rich part relationships on the given shape templates in the learning procedure. Recognizing a chair base as a swivel base significantly reduces the answer space for detecting more fine-grained parts such as central supporting bars, star-base legs and wheels. Inversely, the absence of the chair back strengthens the belief of the recognition of a stool. Different from Sec~\ref{sec:fine_grained_sem_seg} where we consider every segmentation level individually, hierarchical segmentation requires a holistic consideration on the entire part hierarchy.
A key problem towards hierarchical segmentation is how to leverage the rich part relationships on the given shape templates in the learning procedure. 
Recognizing a chair base as a finer-level swivel base significantly reduces the solution space for detecting more fine-grained parts such as central supporting bars, star-base legs and wheels. 
On the other hand, the lack of a chair back increases the possibility that the object is a stool. 
Different from Sec.~\ref{sec:fine_grained_sem_seg} where we consider the problem at each segmentation level separately, hierarchical segmentation requires a holistic understanding on the entire part hierarchy.

{\flushleft \bf Benchmark Algorithms.}
%\paragraph{Benchmark Algorithms.}
% We propose three baseline methods to tackle hierarchical segmentation: bottom-up, top-down and ensemble. The bottom-up method only considers the leaf-node parts during training and groups the prediction of children nodes for parent-node parts along the hierarchies in a bottom-up manner in the inference time. The top-down method learns a multi-labeling task over all the part semantic labels on the tree and conducts top-down inference that we first classify points according to the coarse level semantics and then consider the following fine-grained levels. For the ensemble method, we train flat segmentation at multiple levels as defined in Sec~\ref{sec:fine_grained_sem_seg} and conducts joint inference by calculating the average log likelihood scores over all root-to-leaf paths on the tree. We use PointNet++ \cite{qi2017pointnet++} as the backbone network for all the three methods\footnote{One can also use any of other methods in Sec~\ref{sec:fine_grained_sem_seg} as the backbone.}. More architecture and training details are described in the supplemental material. 
We propose three baseline methods to tackle hierarchical segmentation: bottom-up, top-down and ensemble. 
The bottom-up method considers only the leaf-node parts during training and groups the prediction of the children nodes to parent nodes as defined in the hierarchies in the bottom-up inference. 
The top-down method learns a multi-labeling task over all part semantic labels on the tree and conducts a top-down inference by classifying coarser-level nodes first and then finer-level ones. 
For the ensemble method, we train flat segmentation at multiple levels as defined in Sec.~\ref{sec:fine_grained_sem_seg} and conduct joint inference by calculating the average log-likelihood scores over all the root-to-leaf paths on the tree. 
We use PointNet++~\cite{qi2017pointnet++} as the backbone network in this work, and other methods listed in Sec.~\ref{sec:fine_grained_sem_seg} can also be used. 
More architecture and training details are described in the supplementary material. 
%
%We propose a multi-labeling network training mechanism that learns to perform hierarchical segmentation. Each point in the point cloud may have more than one semantic label when tracing along the path from the root node to the leaf node in the hierarchy. In the training, we learn to predict a confidence score for the point belonging to each semantic label in the tree. Instead of using the SoftMax mutual exclusive activation layer, we use Sigmoid functions to predict confidence scores between $[0, 1]$ and train for each semantic label separately. At inference time, we can either directly predict the labels at each segmentation level or apply parent-child constraints in the hierarchy in a top-down manner.

{\flushleft \bf Evaluation and Results.}
%\paragraph{Evaluation and Results.}
% Table \ref{tab:hier_seg_per_part_category_result} reports the performances for the three methods. We calculate mean IoU for each part category and compute the average over all the tree nodes as the evaluation metric. We observe that the three methods have very small performance gaps. The ensemble method performs better over the other two, especially for the categories with rich structural variation or sub-categorization, such as chair, table and clock.
Table~\ref{tab:hier_seg_per_part_category_result} demonstrates the performances of the three baseline methods. We calculate mIoU for each part category and compute the average over all the tree nodes as the evaluation metric. 
The experimental results show that the three methods perform similarly with small performance gaps. 
The ensemble method performs slightly better over the other two, especially for the categories with rich structural and sub-categorization variation, such as chair, table and clock.

% The bottom-up method only considers leaf-node parts in the training. Though not directly using the template structure, the parent-node semantics for each leaf node part are implicitly encoded in the leaf-node part definitions. For example, the vertical bars for chair back and chair arm are two different leaf nodes in the training. The top-down method explicitly leverages the tree structures in both training and testing time. However, the greedy top-down inference accumulate the prediction errors. The ensemble methods decouples the hierarchical segmentation task at many levels and performs joint inference that taking the predictions at all levels into consideration. Though demonstrating better performances, it trains multiple networks and thus requires more hyper-parameters and longer training time.
The bottom-up method only considers leaf-node parts in the training. Although the template structure is not directly used, the parent-node semantics of leaf nodes are implicitly encoded in the leaf-node part definitions. 
For example, the vertical bars for chair backs and chair arms are two different leaf nodes. 
The top-down method explicitly leverages the tree structures in both the training and the testing phases. 
However, prediction errors are accumulated through top-down inference. 
The ensemble method decouples the hierarchical segmentation task into individual tasks at multiple levels and performs joint inference, taking the predictions at all levels into consideration. 
Though demonstrating better performances, it has more hyper-parameters and requires longer training time for the multiple networks.

\subsection{Instance Segmentation}
\label{sec:ins_seg}

\begin{table*}[th!]
    \footnotesize
    \centering
    \begin{tabular}[width=\linewidth]{@{\hspace{0.1cm}}p{0.3cm}|@{\hspace{0.1cm}}p{0.35cm}|@{\hspace{0.1cm}}p{0.15cm}p{0.15cm}p{0.2cm}p{0.25cm} p{0.35cm}p{0.35cm}p{0.25cm}p{0.25cm} p{0.25cm}p{0.15cm}p{0.25cm}p{0.15cm} p{0.15cm}p{0.35cm}p{0.35cm}p{0.15cm} p{0.35cm}p{0.25cm}p{0.25cm}p{0.15cm} p{0.35cm}p{0.35cm}p{0.35cm}p{0.35cm}}
    \Xhline{2\arrayrulewidth}
     & \textbf{Avg} & \textbf{Bag} & \textbf{Bed} & \textbf{Bott} & \textbf{Bowl} & \textbf{Chair} & \textbf{Clock} & \textbf{Dish} & \textbf{Disp} & \textbf{Door} & \textbf{Ear} & \textbf{Fauc} & \textbf{Hat} & \textbf{Key} & \textbf{Knife} & \textbf{Lamp} & \textbf{Lap} & \textbf{Micro} & \textbf{Mug} & \textbf{Frid} & \textbf{Scis} & \textbf{Stora} & \textbf{Table} & \textbf{Trash} & \textbf{Vase} \\ \Xhline{2\arrayrulewidth}
     \textbf{S1} & 55.7 & 38.8 & 29.8 & 61.9 & 56.9 & 72.4 & 20.3 & 72.2 & 89.3 & 49.0 & \textbf{57.8} & 63.2 & 68.7 & 20.0 & \textbf{63.2} & 32.7 & \textbf{100} & \textbf{50.6} & 82.2 & \textbf{50.6} & 71.7 & 32.9 & 49.2 & 56.8 & 46.6 \\
     \textbf{S2} & 29.7 & $\;\;-$ & 15.4 & $\;\;-$ & $\;\;-$ & 25.4 & $\;\;-$ & 58.1 & $\;\;-$ & 25.4 & $\;\;-$ & $\;\;-$ & $\;\;-$ & $\;\;-$ & $\;\;-$ & 21.7 & $\;\;-$ & \textbf{49.4} & $\;\;-$ & 22.1 & $\;\;-$ & 30.5 & 18.9 & $\;\;-$ & $\;\;-$ \\
     \textbf{S3} & 29.5 & $\;\;-$ & 11.8 & 45.1 & $\;\;-$ & 19.4 & 18.2 & 38.3 & 78.8 & 15.4 & 35.9 & 37.8 & $\;\;-$ & $\;\;-$ & 38.3 & 14.4 & $\;\;-$ & 32.7 & $\;\;-$ & 18.2 & $\;\;-$ & 21.5 & 14.6 & 24.9 & 36.5 \\ \hline
     \textbf{Avg} & 46.8 & 38.8 & 19.0 & 53.5 & 56.9 & 39.1 & 19.3 & 56.2 & 84.0 & 29.9 & 46.9 & 50.5 & 68.7 & 20.0 & 50.7 & 22.9 & \textbf{100} & \textbf{44.2} & 82.2 & 30.3 & 71.7 & 28.3 & 27.5 & 40.9 & 41.6 \\ \Xhline{2\arrayrulewidth}
     \textbf{O1} & \textbf{62.6} & \textbf{64.7} & \textbf{48.4}& \textbf{63.6} & \textbf{59.7} & \textbf{74.4} & \textbf{42.8} & \textbf{76.3} & \textbf{93.3} & \textbf{52.9} & 57.7 & \textbf{69.6} & \textbf{70.9} & \textbf{43.9} & 58.4 & \textbf{37.2} & \textbf{100} & 50.0 & \textbf{86.0} & 50.0 & \textbf{80.9} & \textbf{45.2} & \textbf{54.2} & \textbf{71.7} & \textbf{49.8} \\
     \textbf{O2} & \textbf{37.4} & $\;\;-$ & \textbf{23.0} & $\;\;-$ & $\;\;-$ & \textbf{35.5} & $\;\;-$ & \textbf{62.8} & $\;\;-$ & \textbf{39.7} & $\;\;-$ & $\;\;-$ & $\;\;-$ & $\;\;-$ & $\;\;-$ & \textbf{26.9} & $\;\;-$ & 47.8 & $\;\;-$ & \textbf{35.2} & $\;\;-$ & \textbf{35.0} & \textbf{31.0} & $\;\;-$ & $\;\;-$ \\
     \textbf{O3} & \textbf{36.6} & $\;\;-$ & \textbf{15.0} & \textbf{48.6} & $\;\;-$ & \textbf{29.0} & \textbf{32.3} & \textbf{53.3} & \textbf{80.1} & \textbf{17.2} & \textbf{39.4} & \textbf{44.7} & $\;\;-$ & $\;\;-$ & \textbf{45.8} & \textbf{18.7} & $\;\;-$ & \textbf{34.8} & $\;\;-$ & \textbf{26.5} & $\;\;-$ & \textbf{27.5} & \textbf{23.9} & \textbf{33.7} & \textbf{52.0} \\ \hline
     \textbf{Avg} & \textbf{54.4} & \textbf{64.7} & \textbf{28.8} & \textbf{56.1} & \textbf{59.7} & \textbf{46.3} & \textbf{37.5} & \textbf{64.1} & \textbf{86.7} & \textbf{36.6} & \textbf{48.5} & \textbf{57.1} & \textbf{70.9} & \textbf{43.9} & \textbf{52.1} & \textbf{27.6} & \textbf{100} & \textbf{44.2} & \textbf{86.0} & \textbf{37.2} & \textbf{80.9} & \textbf{35.9} & \textbf{36.4} & \textbf{52.7} & \textbf{50.9} \\ \Xhline{2\arrayrulewidth}
    \end{tabular}
    \vspace{-1mm}
    \caption{\textbf{Instance segmentation results (part-category mAP \%, IoU threshold 0.5).} Algorithm \textbf{S} and \textbf{O} refer to SGPN~\cite{wang2018sgpn} and our proposed method respectively. The number \textbf{1}, \textbf{2} and \textbf{3} refer to the three levels of segmentation: coarse-, middle- and fine-grained. We put short lines for the levels that are not defined.}
    \label{tab:ins_seg_per_part_category_result}
    \vspace{-3mm}
\end{table*}

% The goal of instance segmentation is to detect every individual part instance and segment it out from the context of the shape. Many applications in the fields of compute graphics, vision and robotics, including manufacturing, assembly, interaction and manipulation, require the instance-level part recognition. Compared to detecting objects from scenes, parts on objects usually have stronger and more intertwined structural relationships than scenes. The existence of many visually similar parts with different semantics further complicates the part detection problem. To the best of our knowledge, PartNet provides the first large-scale object instance segmentation benchmark.
The goal of instance segmentation is to detect every individual part instance and segment it out from the context of the shape. 
Many applications in computer graphics, vision and robotics, including manufacturing, assembly, interaction and manipulation, require the instance-level part recognition. 
Compared to detecting objects from scenes, parts on objects usually have stronger and more intertwined structural relationships. 
The existence of many visually-similar but semantically-different parts makes the part detection problem challenging. 
To the best of our knowledge, this work is the first to provide a large-scale shape part instance-level segmentation benchmark.

% Given a category of shape point clouds as input, the task of shape part instance segmentation is to provide several disjoint masks over the entire point clouds, each of which corresponds to a detected instance of part with a semantic label within a pool of parts of interests. We adapt the part semantics from the multiple levels of semantic segmentation that we use in Sec~\ref{sec:fine_grained_sem_seg}. The detected masks should have no overlaps, but unnecessarily cover the entire point clouds, due to the existence of points with no semantic labels.
Given a shape point cloud as input, the task of part instance segmentation is to provide several disjoint masks over the entire point cloud, each of which corresponds to an individual part instance on the object. 
We adopt the part semantics from the defined segmentation levels in Sec.~\ref{sec:fine_grained_sem_seg}. 
The detected masks should have no overlaps, but they together do not necessarily cover the entire point cloud, as some points may not belong to any part of interests.

\begin{figure}[t!]
    \centering
    \includegraphics[width=\linewidth]{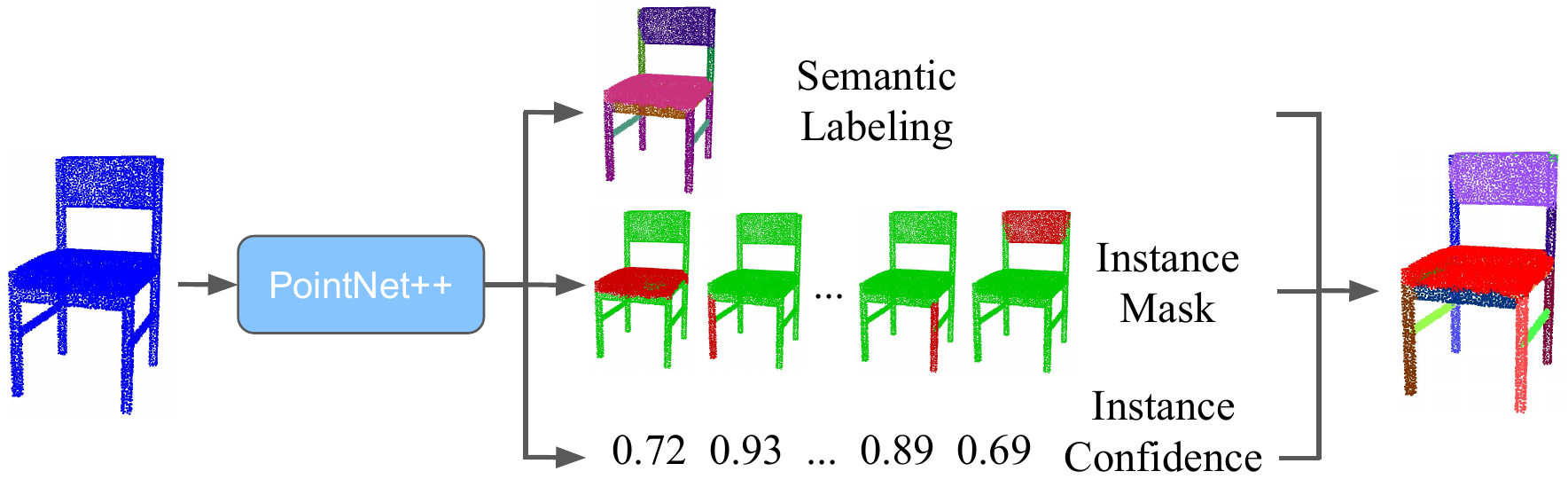}
    \caption{\textbf{The proposed detection-by-segmentation method for instance segmentation.} The network learns to predict three components: the semantic label for each point, a set of disjoint instance masks and their confidence scores for part instances.}
    \label{fig:ins_seg_arch}
    \vspace{-3mm}
\end{figure}

{\flushleft \bf Benchmark Algorithms.}
%\paragraph{Benchmark Algorithms.}
% We propose a detection-by-segmentation network to address instance segmentation. We illustrate our network architecture in Figure~\ref{fig:ins_seg_arch}. We use PointNet++ \cite{qi2017pointnet++} as the backbone network for feature extraction and learn to predict both semantic segmentation for each point and $K$ instance segmentation masks $\{\hat{y}_i\in [0,1]^{N}|i=1,2,\cdots, K\}$ over the input point clouds of size $N$. We also train a separate mask $\hat{y}_{other}$ for all the points with no semantic labels in the ground-truth. A softmax activation layer is applied to encourage the mutual exclusiveness among different masks such that $\hat{y}_1+\hat{y}_2+\cdots+\hat{y}_K+\hat{y}_{other}=\mathbf{1}$. To train the network, we run Hungarian algorithm \cite{kuhn1955hungarian} to find a bipartite matching $\mathcal{M}: \left\{i\rightarrow \mathcal{M}(i)|i=1,2,\cdots,T\right\}$ between the prediction masks $\{\hat{y}_i|i=1,2,\cdots, K\}$ and the ground-true masks $\{y_i|i=1,2,\cdots,T\}$ and regress each matched prediction $\hat{y}_{\mathcal{M}(t)}$ to the ground-truth mask $y_t$. We employ a relaxed version of the Intersection over Union (IoU) score \cite{krahenbuhl2013parameter} defined as $IoU(p, q) = \langle p, q\rangle / (\lVert p\rVert_1+\lVert q\rVert_1-\langle p, q\rangle)$ as the matching scores for Hungarian algorithm. We also train a separate branch to predict confidence scores for the existence of the predicted masks $\left\{C_i|i=1,2,\cdots,K\right\}$.
We propose a novel detection-by-segmentation network to address instance segmentation. We illustrate our network architecture in Figure~\ref{fig:ins_seg_arch}.
We use PointNet++~\cite{qi2017pointnet++} as the backbone network for extracting features and predicting both semantic segmentation for each point and $K$ instance segmentation masks $\{\hat{y}_i\in [0,1]^{N}|i=1,2,\cdots, K\}$ over the input point cloud of size $N$. 
Moreover, we train a separate mask $\hat{y}_{other}$ for the points without semantic labels in the ground-truth. 
A softmax activation layer is applied to encourage the mutual exclusiveness among different masks so that $\hat{y}_1+\hat{y}_2+\cdots+\hat{y}_K+\hat{y}_{other}=\mathbf{1}$. 
To train the network, we apply Hungarian algorithm~\cite{kuhn1955hungarian} to find a bipartite matching $\mathcal{M}: \left\{i\rightarrow \mathcal{M}(i)|i=1,2,\cdots,T\right\}$ between the prediction masks $\{\hat{y}_i|i=1,2,\cdots, K\}$ and the ground-true masks $\{y_i|i=1,2,\cdots,T\}$, and regress each prediction $\hat{y}_{\mathcal{M}(t)}$ to the matched ground-truth mask $y_t$. 
We employ a relaxed version of IoU~\cite{krahenbuhl2013parameter} defined as $\text{IoU}(p, q) = \langle p, q\rangle / (\lVert p\rVert_1+\lVert q\rVert_1-\langle p, q\rangle)$, as the metric for Hungarian algorithm. 
In the meanwhile, a separate branch is trained to predict confidence scores for the predicted masks $\left\{C_i|i=1,2,\cdots,K\right\}$.

% The training loss $L=L_{sem}+\lambda_{ins}L_{ins}+\lambda_{other}L_{other}+\lambda_{conf}L_{conf}+\lambda_{l21}L_{l21}$ combines five terms: the regular cross-entry semantic segmentation loss $L_{sem}$, the relaxed IoU loss for each matched mask prediction $L_{ins}=\sum_{i=1}^T IoU(\hat{y}_{\mathcal{M}(i)}, y_i)$, the related IoU loss for the unlabeled point $L_{other}=IoU(\hat{y}_{other}, y_{other})$, the confidence prediction loss $L_{conf}=\sum_{i=1}^T (C_{\mathcal{M}(i)}-IoU(\hat{y}_{\mathcal{M}(i)}, y_i))^2$ and a $l_{2,1}$-norm regularization term $L_{l21}=\sum_{i=1}^K\lVert \hat{y}_i \rVert_2^2 + \lVert \hat{y}_{other}\rVert_2^2$ that is shown in \cite{sung2018deep} to encourage structural sparsity of the prediction masks. We use $N=10,000$, $K=200$, $\lambda_{ins}=1.0$, $\lambda_{other}=1.0$, $\lambda_{conf}=1.0$ and $\lambda_{l21}=0.1$ for all experiments.
The loss function is defined as $L=L_{sem}+\lambda_{ins}L_{ins}+\lambda_{other}L_{other}+\lambda_{conf}L_{conf}+\lambda_{l21}L_{l21}$,  combining five terms: a cross-entropy semantic segmentation loss $L_{sem}$, an IoU loss for mask regression $L_{ins}=\sum_{i=1}^T \text{IoU}(\hat{y}_{\mathcal{M}(i)}, y_i)$, an IoU loss for the unlabeled points $L_{other}=\text{IoU}(\hat{y}_{other}, y_{other})$, a prediction-confidence loss $L_{conf}=\sum_{i=1}^T (C_{\mathcal{M}(i)}-\text{IoU}(\hat{y}_{\mathcal{M}(i)}, y_i))^2$ and a $l_{2,1}$-norm regularization term $L_{l21}=\sum_{i=1}^K\lVert \hat{y}_i \rVert_2 + \lVert \hat{y}_{other}\rVert_2$ to encourage unused prediction masks to vanish~\cite{sung2018deep}. 
We use $N=10,000$, $K=200$, $\lambda_{ins}=1.0$, $\lambda_{other}=1.0$, $\lambda_{conf}=1.0$ and $\lambda_{l21}=0.1$ for all the experiments.

% We compare the proposed method to SGPN \cite{wang2018sgpn}, which learns similarity scores among all pairs of points and detect part instances by grouping points sharing similar features. We follow most of the default settings and hyper-parameters in the paper. We first pretrain PointNet++ semantic segmentation branch and then finetune to train for the per-point feature similarity matrix and confidence maps. We input 10,000 points to the network and use margin values of 1 and 2 for the double-hinge loss\footnote{In the original paper, the margin values are 10 and 80. The authors of SGPN suggest that using 1 and 2 gives better results.}. We use batch-size 32 during the pretraining and batch-size 1 during the finetuning.
We compare the proposed method with SGPN~\cite{wang2018sgpn}, which learns similarity scores among all pairs of points and detect part instances by grouping points that share similar features. 
We follow most of the default settings and hyper-parameters described in their paper. 
We first pre-train PointNet++ semantic segmentation branch and then fine-tune it for improving the per-point feature similarity matrix and confidence maps. 
We use margin values of 1 and 2 for the double-hinge loss as suggested by the authors of~\cite{wang2018sgpn}, instead of 10 and 80 in the original paper.
We feed 10,000 points to the network at a time, and use a batch-size of 32 in the pre-training and 1 in the fine-tuning.

\begin{figure}[t]
    \centering
    \includegraphics[width=0.8\linewidth]{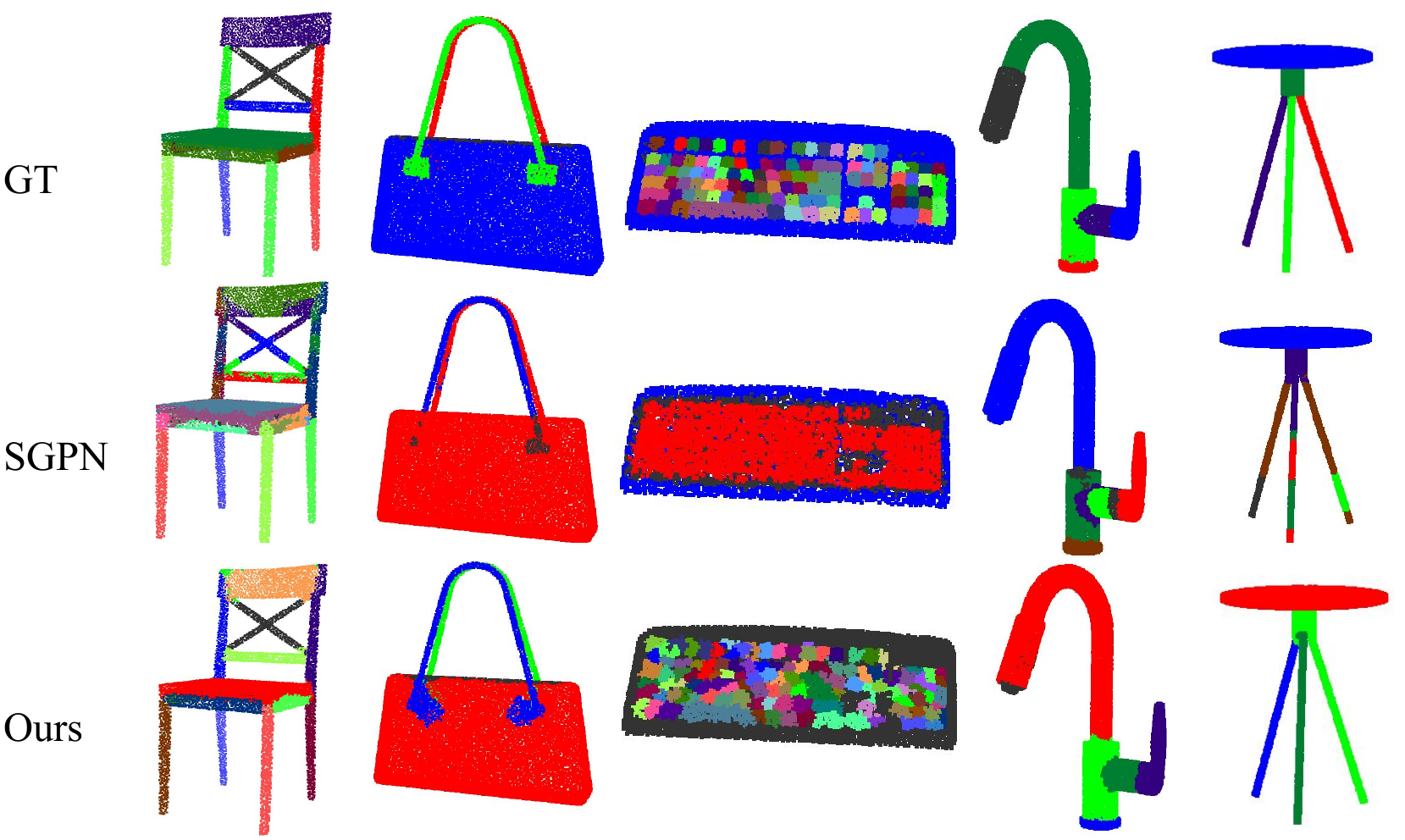}
    \caption{\textbf{Qualitative results for instance segmentation.} Our method produces more robust and cleaner results than SGPN.}
    \label{fig:ins_seg_result}
    \vspace{-3mm}
\end{figure}

\begin{figure}[t]
    \centering
    \includegraphics[width=0.8\linewidth]{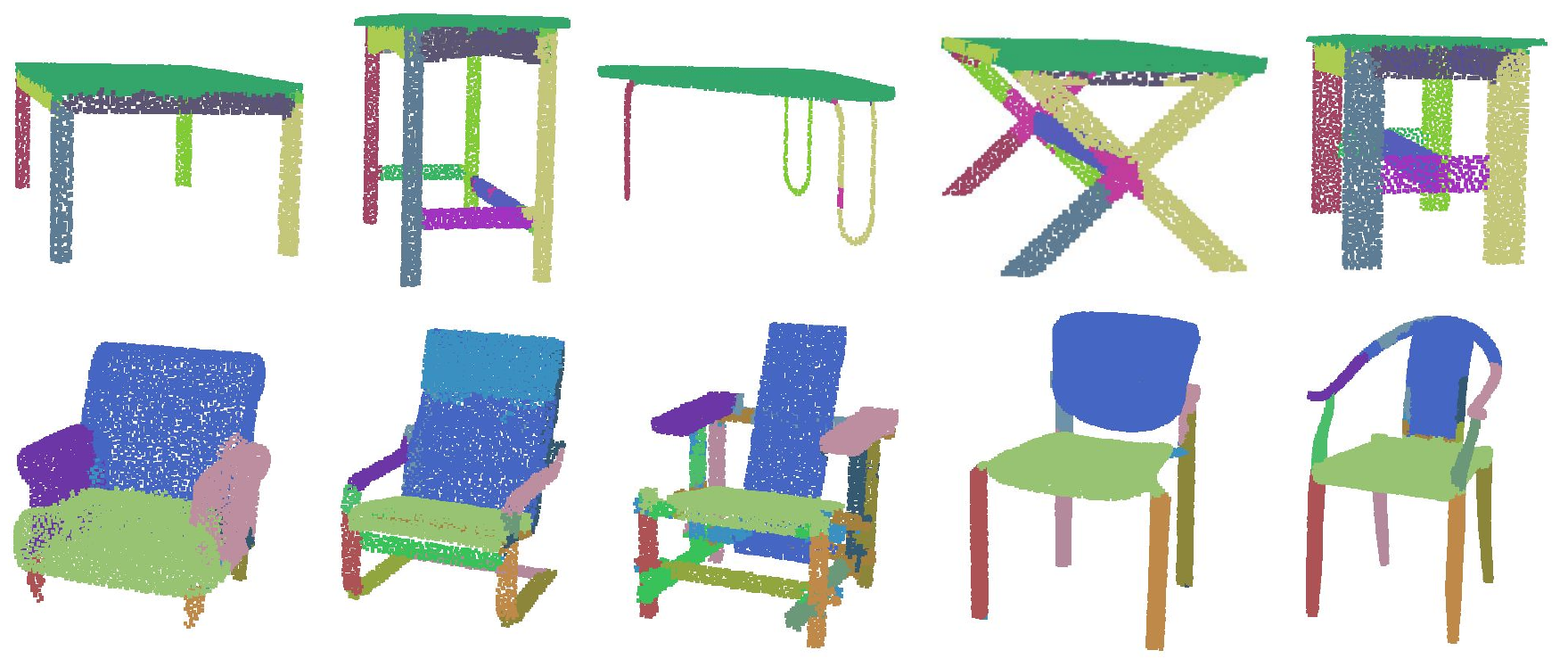}
    \caption{\textbf{Learned instance correspondences.} The corresponding parts are marked with the same color.}
    \label{fig:ins_seg_cor}
    \vspace{-3mm}
\end{figure}

{\flushleft \bf Evaluation and Results.}
%\paragraph{Evaluation and Results.}
Table~\ref{tab:ins_seg_per_part_category_result} reports the per-category mean Average Precision (mAP) scores for SPGN and our proposed method. For each object category, the mAP score calculates the AP for each semantic part category across all test shapes and averages the AP across all part categories. Finally, we take the average of the mAP scores across different levels of segmentation within each object category and then report the average over all object categories. We compute the IoU between each prediction mask and the closest ground-truth mask and regard a prediction mask as true positive when IoU is larger than $0.5$. 
% Moved to supplemental material: We also report the per-shape mAP in Table \ref{tab:ins_seg_per_shape_result}, where we calculate mAP for the part categories that either occur in the prediction or in the ground-truth within each shape and averages the scores among all test shapes. The per-category mAP equally balances all part categories, while the per-shape mAP reports the on-average instance segmentation performance on a randomly given shape.

Figure~\ref{fig:ins_seg_result} shows qualitative comparisons for our proposed method and SGPN. Our proposed method produces more robust and cleaner instance predictions. After learning for point features, SGPN has a post-processing stage that merges points with similar features as one component. This process involves many thresholding hyper-parameters. Even though most parameters are automatically inferred from the validation data, SPGN still suffers from predicting partial or noisy instances in the case of bad thresholding. Our proposed method learns structural priors within each object category that is more instance-aware and robust in predicting complete instances. We observe that training for a set of disjoint masks across multiple shapes gives us consistent part instances. We show the learned part correspondence in Figure~\ref{fig:ins_seg_cor}.
%Qualitative comparisons between our proposed method and SGPN are provided in Figure~\ref{fig:ins_seg_result}. 
%Overall, the proposed method produces more robust and cleaner instance predictions, which can be seen from the examples of keyboard and table. 
%In SGPN, a post-processing stage has been used to merges point features with similar features in one component. 
%This process involves a few thresholding with fixed hyper-parameters, which hardly generalize to all the scenarios. 
%On the other hand, the proposed method learns instance-aware structural priors within each object category, which helps predicting complete instances robustly. 
%Similar to \cite{sung2018deep}, We show the learned part correspondence in Figure~\ref{fig:ins_seg_cor}, to illustrate part instance consistency across multiple shapes. 

%% file: tex/conclusion.tex
We introduce PartNet: a \emph{large-scale} benchmark for \emph{fine-grained, hierarchical, and instance-level} 3D shape segmentation. 
It contains $573,585$ part annotations for $26,671$ ShapeNet \cite{chang2015shapenet} models from $24$ object categories. 
%We design a 3D annotation interface with question-answering workflow to collect the fine-grained and hierarchical instance-level shape part decomposition. 
Based on the dataset, we propose three shape segmentation benchmarks: fine-grained semantic segmentation, hierarchical semantic segmentation and instance segmentation. 
We benchmark four state-of-the-art algorithms for semantic segmentation and propose a novel method for instance segmentation that outperforms the existing baseline method. 
Our dataset enables future research directions such as collecting more geometric and semantic annotation on parts, investigating shape grammars for synthesis and animating object articulation in virtual environments for robotic learning.

%% file: tex/ack.tex
This research was supported by NSF grants CRI-1729205 and IIS-1763268, a Vannevar Bush Faculty Fellowship, a Google fellowship, and gifts from Autodesk, Google and Intel AI Lab.
We especially thank Zhe Hu from Hikvision for the help on data annotation and Linfeng Zhao for the help on preparing hierarchical templates.
We appreciate the 66 annotators from Hikvision, Ytuuu and Data++ on data annotation.

%% file: tex_supp/data.tex
We present more visualization and statistics over the proposed PartNet dataset. 

\subsection{More Fine-grained Segmentation Visualization}
Figure~\ref{fig:fine_grained_visu1} and \ref{fig:fine_grained_visu2} show more visualization for  fine-grained instance-level segmentation annotations in PartNet. We observe the complexity of the annotated segmentation and the heterogeneous variation of shapes within each object category.

\subsection{More Hierarchical Segmentation Visualization}
Figure~\ref{fig:hier_visu1}, \ref{fig:hier_visu2} and \ref{fig:hier_visu3} show more visualization for example hierarchical instance-level segmentation annotations in PartNet. We visualize the tree-structure of the hierarchical segmentation annotation with the 2D part renderings associated to the tree nodes.

\subsection{Shape Statistics}
We report the statistics for the number of annotations, unique shapes and shapes that we collect multiple human annotations in Figure~\ref{fig:shape_histogram}.

\subsection{Part Statistics}
We report the statistics for the number of part semantics for each object category in Figure~\ref{fig:part_sem}. We also present the statistics for the maximum and median number of part instances per shape for each object category in Figure~\ref{fig:part_count}. We report the statistics for the maximum and median tree depth for each object category in Figure~\ref{fig:tree_depth}.

%% file: tex_supp/template.tex
We provide more details and visualization for the expert-defined hierarchical templates to guide the hierarchical segmentation annotation and the template refinement procedure to resolve annotation inconsistencies.

\begin{figure}[t!]
\centering
\includegraphics[width=\linewidth]{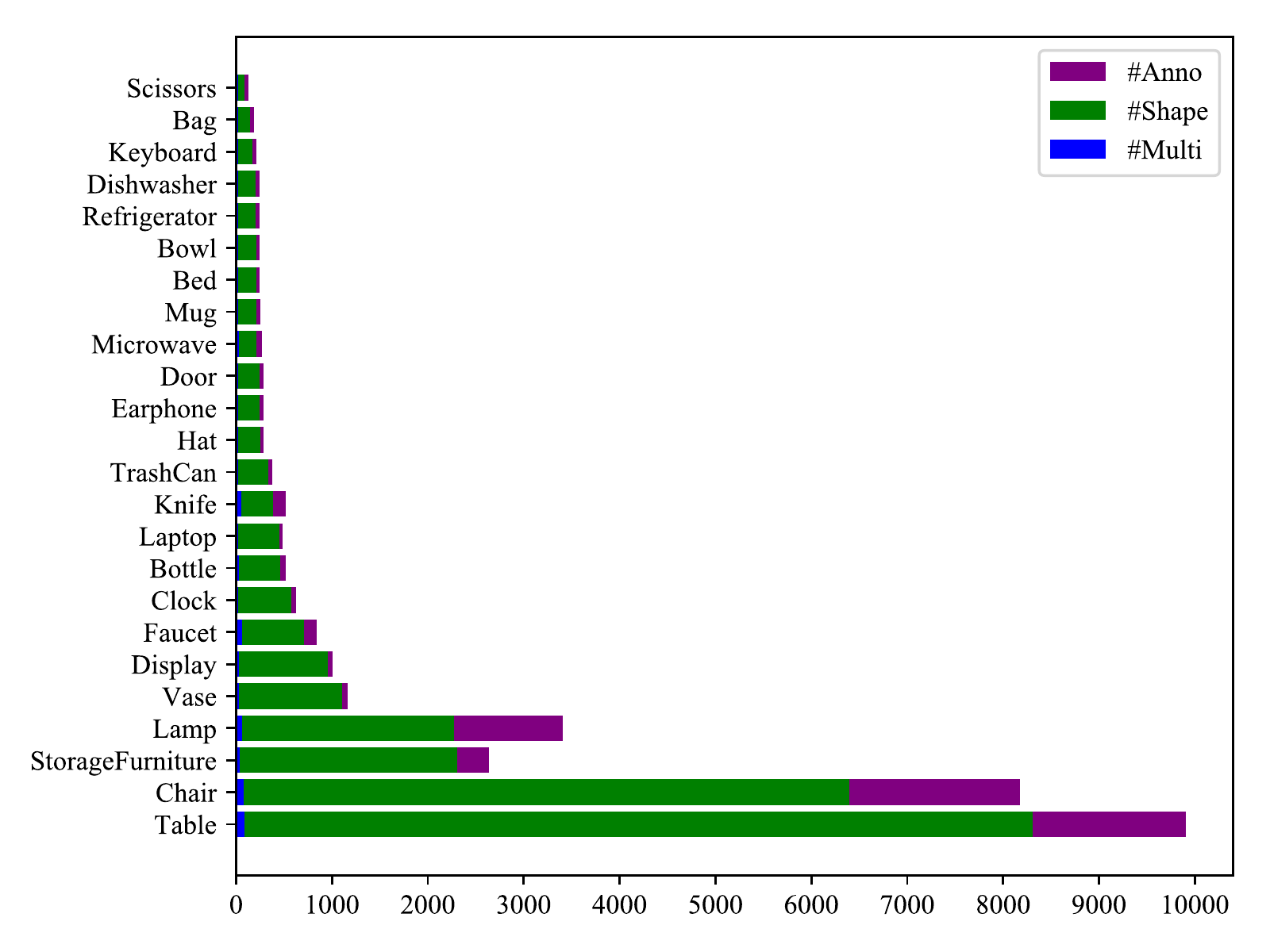}
\caption{\textbf{PartNet shape statistics.} We report the statistics for the number of annotations, unique shapes and shapes that we collect multiple human annotations.}
\label{fig:shape_histogram}
\end{figure}

\begin{figure}[t!]
\centering
\includegraphics[width=\linewidth]{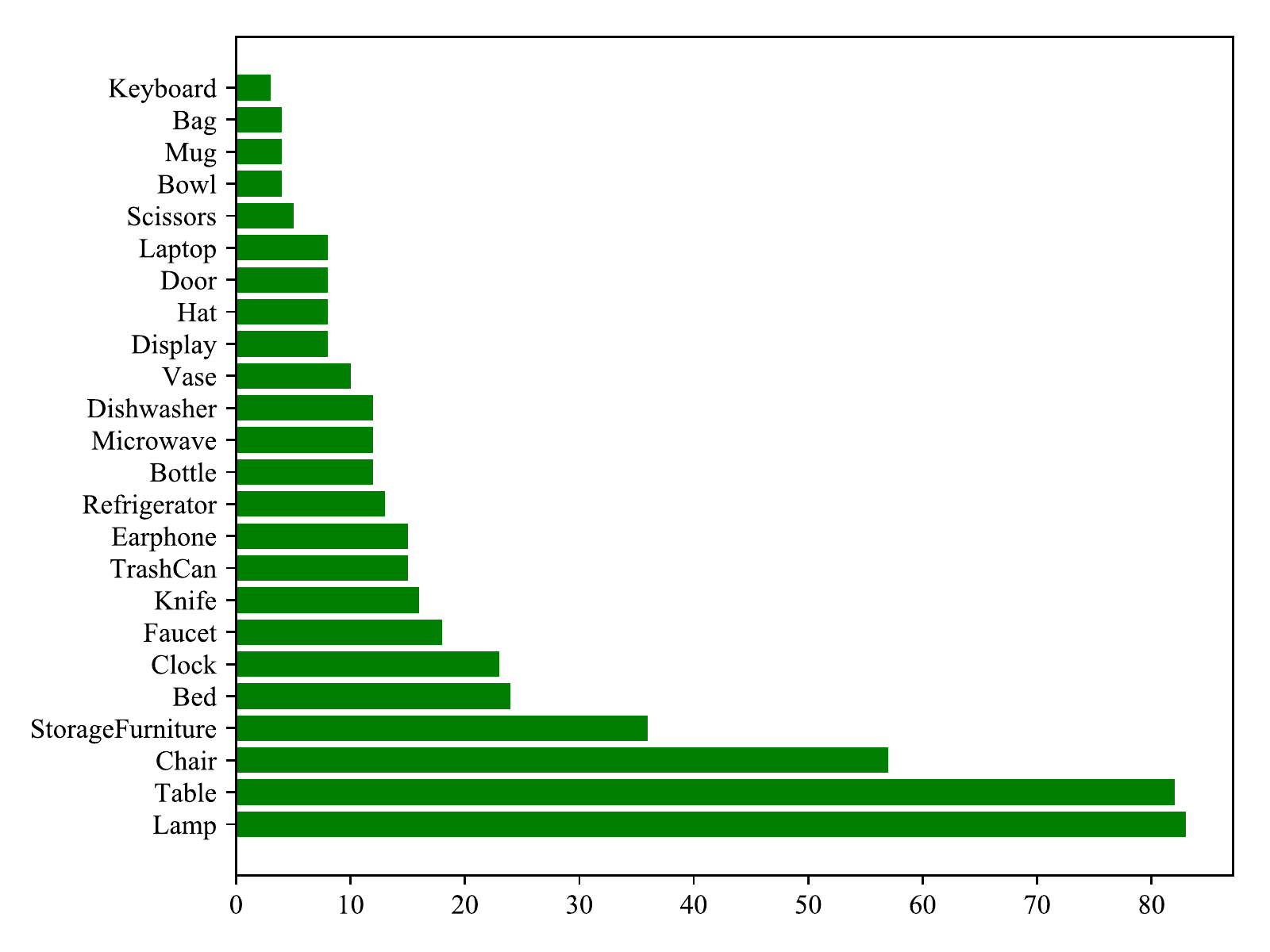}
\caption{\textbf{PartNet part semantics statistics.} We report the statistics for the number of part semantics for each object category.}
\label{fig:part_sem}
\end{figure}

\begin{figure}[t!]
\centering
\includegraphics[width=\linewidth]{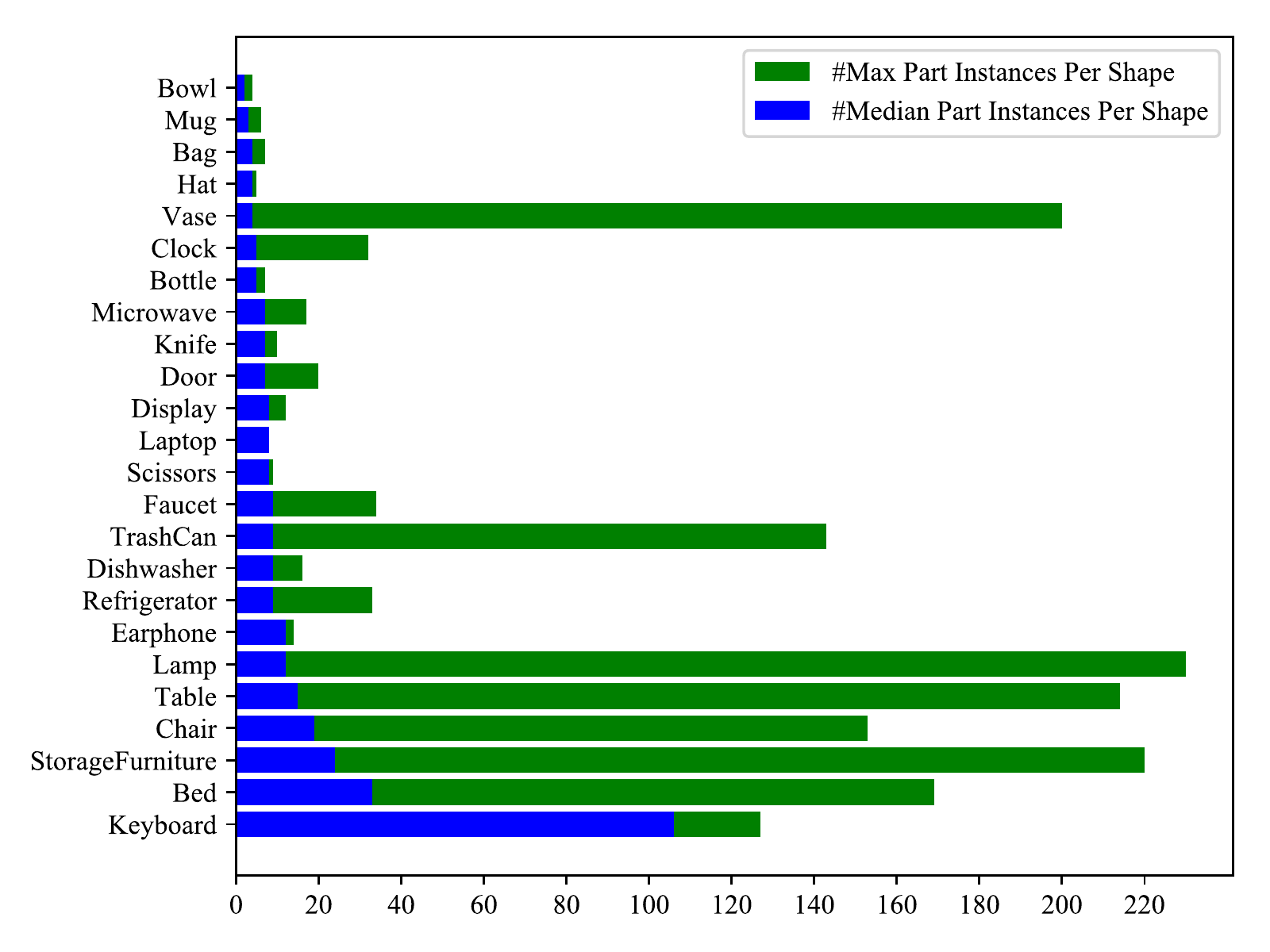}
\caption{\textbf{PartNet part instance statistics.} We report the statistics for the maximum and median number of part instances per shape for each object category.}
\label{fig:part_count}
\end{figure}

\begin{figure}[t!]
\centering
\includegraphics[width=\linewidth]{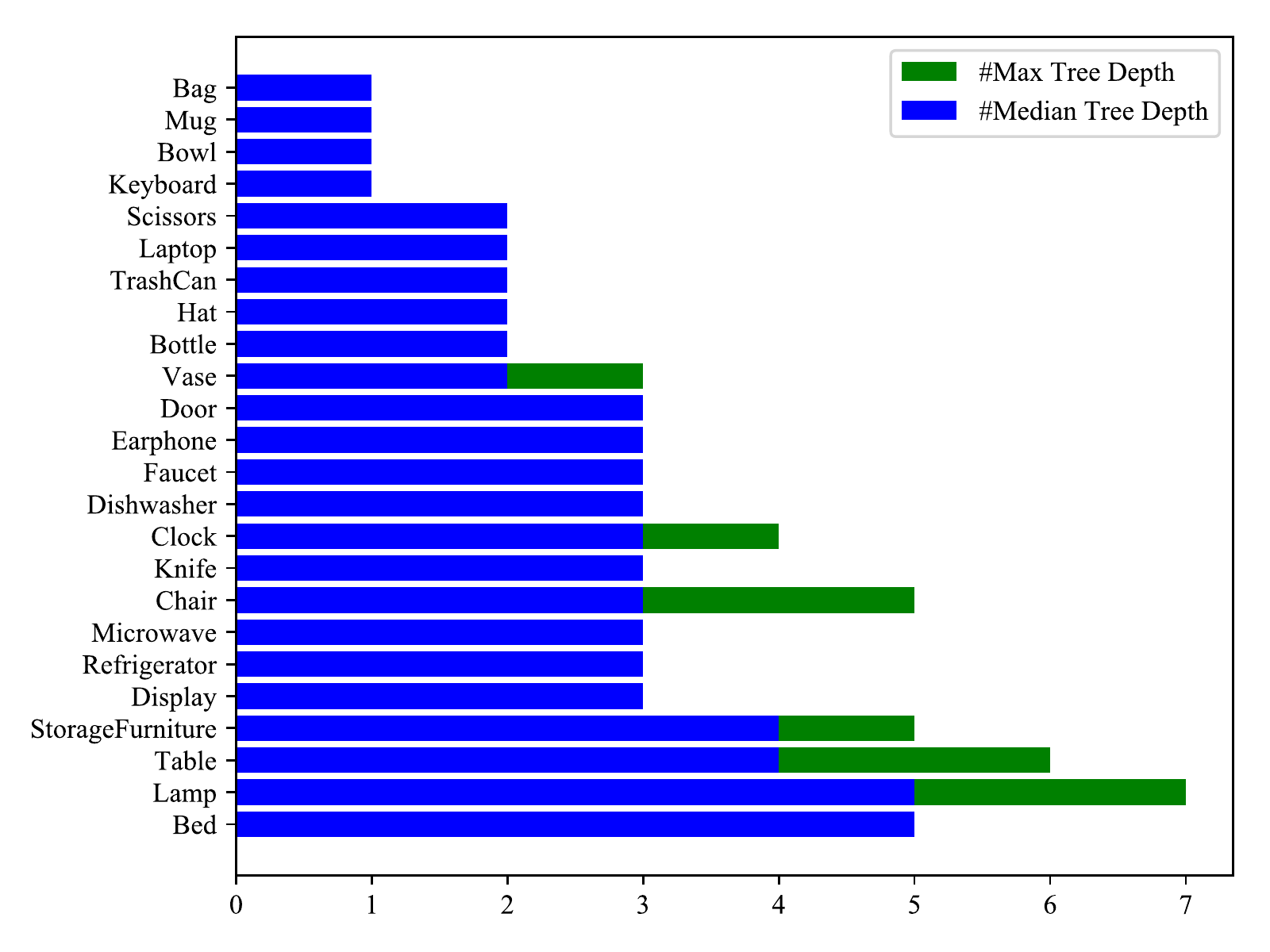}
\caption{\textbf{PartNet tree depth statistics.} We report the statistics for the maximum and median tree depth for each object category.}
\label{fig:tree_depth}
\end{figure}

\subsection{Template Design Details}

We design templates according to the rules of thumb that we describe in the main paper. We also consulted many online references\footnote{\Eg \href{http://www.props.eric-hart.com/resources/parts-of-a-chair/}{http://www.props.eric-hart.com/resources/parts-of-a-chair/}.} that describe object parts (often for manufacturing and assembly) and previous works that relate language to the shapes~\cite{chang2018linking} as guides for the design of our template.  To ensure that our templates cover most of the shape variations and part semantics of each object category, we generated a t-SNE~\cite{maaten2008visualizing} visualization for the entire shape space to study the shape variation.   We trained an auto-encoder based on the shape geometry within each object category to obtain shape embeddings for the t-SNE visualization.

\begin{table*}[th!]
    \footnotesize
    \centering
    \begin{tabular}[width=\linewidth]{@{\hspace{0.cm}}p{0.35cm}|@{\hspace{0.05cm}}p{0.3cm}|@{\hspace{0.05cm}}p{0.15cm}p{0.15cm}p{0.2cm}p{0.25cm} p{0.35cm}p{0.35cm}p{0.25cm}p{0.25cm} p{0.25cm}p{0.15cm}p{0.25cm}p{0.15cm} p{0.15cm}p{0.35cm}p{0.35cm}p{0.15cm} p{0.35cm}p{0.25cm}p{0.25cm}p{0.15cm} p{0.35cm}p{0.35cm}p{0.35cm}p{0.35cm}}
    \Xhline{2\arrayrulewidth}
     & \textbf{Avg} & \textbf{Bag} & \textbf{Bed} & \textbf{Bott} & \textbf{Bowl} & \textbf{Chair} & \textbf{Clock} & \textbf{Dish} & \textbf{Disp} & \textbf{Door} & \textbf{Ear} & \textbf{Fauc} & \textbf{Hat} & \textbf{Key} & \textbf{Knife} & \textbf{Lamp} & \textbf{Lap} & \textbf{Micro} & \textbf{Mug} & \textbf{Frid} & \textbf{Scis} & \textbf{Stora} & \textbf{Table} & \textbf{Trash} & \textbf{Vase} \\ 
     \Xhline{2\arrayrulewidth}
     \textbf{O\textsubscript{avg}} & 69.8 & 54.8  & 70.0  & 87.5  & 87.7  & 59.0  & 62.1  & 67.3  & 85.2  & 64.4  & 74.1  & 69.9  & 86.8  & 77.0  & 75.0  & 44.6  & 61.6  & 71.0  & 91.0  & 65.3  & 88.7  & 68.0  & 40.1  & 51.4  & 72.6 \\
     \textbf{O\textsubscript{avg}} & 19.0  & 29.3  & 17.4  & 6.9  & 8.1  & 26.7  & 24.6  & 19.1  & 17.8  & 15.2  & 15.6  & 17.6  & 9.5  & 17.9  & 14.6  & 29.0  & 27.1  & 19.3  & 10.6  & 27.7  & 9.0  & 21.3  & 29.3  & 23.3  & 19.5 \\
     \Xhline{2\arrayrulewidth}
     \textbf{R\textsubscript{avg}} & 83.3  & 82.1  & 76.2  & 89.3  & 91.7  & 77.8  & 91.1  & 81.5  & 94.0  & 77.0  & 83.0  & 84.7  & 89.3  & 89.6  & 77.8  & 72.7  & 78.3  & 84.4  & 91.7  & 85.1  & 90.2  & 77.1  & 71.4  & 71.0  & 92.3 \\
     \textbf{R\textsubscript{std}} & 10.4  & 11.1  & 9.2  & 6.0  & 7.4  & 15.2  & 7.0  & 7.2  & 2.8  & 11.2  & 13.5  & 8.6  & 10.3  & 3.5  & 14.9  & 17.3  & 14.6  & 9.1  & 9.7  & 6.2  & 8.6  & 12.8  & 22.6  & 13.2  & 7.5 \\
     \Xhline{2\arrayrulewidth}
    \end{tabular}
    \caption{\textbf{The average confusion scores and the standard deviations for multiple annotations (\%).} We report the average confusion scores and the standard deviations by calculating over the entries on the diagonal of the confusion matrix for each object category using the small subset of shapes that we collect multiple human annotations. Rows \textbf{O} and \textbf{R} respectively refer to the scores before and after the template refinement process.}
    \label{tab:confusion_scores}
\end{table*}

Although we try to cover the most common part semantics in our templates, it is still not easy to cover all possible object parts. Thus, we allow annotators to deviate from the templates and define their own parts and segmentation structures. Among all the annotated part instances, 1.3\% of them are defined by the annotators.   In the raw annotation, 13.1\% of shapes contained  user-defined part labels. 

Our analysis shows that our template designs are able to cover most of the ShapeNet~\cite{chang2015shapenet} shapes. Of the 27,260 shapes we collected in total, our annotators successfully labeled 26,671 of them, giving our templates a coverage rate of at least 97.8\% for ShapeNet shapes. While template coverage is a potential issue, the remaining 2.2\% were not annotated mainly due to other issues such as poor mesh quality, classification error, error during mesh splitting, \textit{etc.}

We design hierarchical templates that cover both the coarse-level part semantics and fine-grained part details down to the primitive level, \eg \textit{chair back vertical bar} and \textit{bed base surface panel}. Most primitive-level parts are atomic such that they are very unlikely to be further divided for end applications. If an application requires different segmentation hierarchy or level of segmentation than the ones we already provide in our template, developers and researchers can try to build up their own segmentation based upon the atomic primitives we obtain in PartNet.

Moreover, we try our best to make the shared part concepts among different shapes and even different object categories share the same part labels. For example, we use the part label \textit{leg} for \textit{table}, \textit{chair}, \textit{lamp base}, \textit{etc.} and the part label \textit{wheel} for both \textit{chair swivel base wheel} and \textit{refrigerator base wheel}. Such part concept sharing provides rich part correspondences within a specific object category and across multiple object categories.

\subsection{Template Refinement Details}

Fine-grained shape segmentation is challenging to annotate due to the subtle concept gaps among similar part semantics.
Even though we provide detailed textual and visual explanation for our pre-defined parts, we still observe some annotation inconsistencies across multiple annotators. To quantitatively diagnose such issues, we reserve a small subset of shapes for which we collect multiple human annotations. Then, we compute the confusion scores among the predefined parts across the multiple annotations and conduct careful template refinement to reduce the part ambiguity.

There are primarily three sources of such inconsistencies: boundary ambiguity, granularity ambiguity and part labeling ambiguity. 
Boundary ambiguity refers to the unclear boundary between two parts, which is also commonly seen in previous works~\cite{chen2009benchmark, yi2016scalable}. For example, the boundary between the bottle neck and the bottle body is not that clear for wine bottles. 
Granularity ambiguity means that different annotators have different understanding about the segmentation granularity of the defined parts. One example is that, for a curvy and continuous chair arm, one can regard it as a whole piece or imagine the separation of armrest and arm support. 
The most common type of ambiguity in our dataset is the part labeling ambiguity. The fine-grained part concepts, though intended to be different category-wise, may apply to the same part on a given object. For example, a connecting structure between the seat and the base of a chair can be considered as chair seat support or chair base connector.

We study the mutual human agreement on the multiple annotation subset. We consider the parts defined at the leaf node level of segmentation on the hierarchy and compute the confusion matrix across multiple human annotations\footnote{We consider the entire path labels as histories to the leaf nodes when computing the confusion matrix.}. The ideal confusion matrix should be close to the diagonal matrix without any part-level ambiguity.
In our analysis, we observe human disagreement among some of our initial part definitions. To address the ambiguity, we either merge two similar concepts with high confusion scores or remove the hard-to-distinguish parts from evaluation. For example, we find our annotators often mix up the annotation for regular tables and desks due to the similarity in the two concepts.  Thus, we merge the desk subtype into the regular table subtype to address this issue. In other cases, some small parts such as the buttons on the displays are very tricky to segment out from the main display frame.  Since they may not be reliably segmented out, we decided to remove such unclear segmentation from evaluation. 

\begin{table*}[th!]
    \footnotesize
    \centering
   \begin{tabular}[width=\linewidth]{@{\hspace{0.1cm}}p{0.3cm}|@{\hspace{0.1cm}}p{0.35cm}|@{\hspace{0.1cm}}p{0.15cm}p{0.15cm}p{0.2cm}p{0.25cm} p{0.35cm}p{0.35cm}p{0.25cm}p{0.25cm} p{0.25cm}p{0.15cm}p{0.25cm}p{0.15cm} p{0.15cm}p{0.35cm}p{0.35cm}p{0.15cm} p{0.35cm}p{0.25cm}p{0.25cm}p{0.15cm} p{0.35cm}p{0.35cm}p{0.35cm}p{0.35cm}}
    \Xhline{2\arrayrulewidth}
     & \textbf{Avg} & \textbf{Bag} & \textbf{Bed} & \textbf{Bott} & \textbf{Bowl} & \textbf{Chair} & \textbf{Clock} & \textbf{Dish} & \textbf{Disp} & \textbf{Door} & \textbf{Ear} & \textbf{Fauc} & \textbf{Hat} & \textbf{Key} & \textbf{Knife} & \textbf{Lamp} & \textbf{Lap} & \textbf{Micro} & \textbf{Mug} & \textbf{Frid} & \textbf{Scis} & \textbf{Stora} & \textbf{Table} & \textbf{Trash} & \textbf{Vase} \\ \Xhline{2\arrayrulewidth}
     \textbf{P1} & 71.8  & 59.3  & 39.6  & 81.0  & 78.5  & 81.8  & 67.1  & 78.9  & 88.2  & 71.1  & 68.0  & 67.5  & 58.5  & 65.6  & 66.5  & 46.5  & 96.5  & 75.0  & 84.2  & \textbf{79.6}  & 86.5  & 55.9  & 85.6  & 66.7  & 76.3  \\
\textbf{P2} & 50.1  & $\;\;-$  & 21.3  & $\;\;-$  & $\;\;-$  & 52.4  & $\;\;-$  & 60.0  & $\;\;-$  & 47.1  & $\;\;-$  & $\;\;-$  & $\;\;-$  & $\;\;-$  & $\;\;-$  & 43.5  & $\;\;-$  & \textbf{64.3}  & $\;\;-$  & \textbf{63.9}  & $\;\;-$  & 48.8  & 50.0  & $\;\;-$  & $\;\;-$  \\
\textbf{P3} & 48.2  & $\;\;-$  & 13.0  & 55.3  & $\;\;-$  & 44.8  & 37.8  & 55.2  & 79.0  & 38.8  & 47.5  & 55.5  & $\;\;-$  & $\;\;-$  & 40.0  & 34.7  & $\;\;-$  & 54.5  & $\;\;-$  & \textbf{53.2}  & $\;\;-$  & 47.4  & 42.5  & 46.4  & 74.0  \\
\hline
\textbf{Avg} & 63.4  & 59.3  & 24.6  & 68.2  & 78.5  & 59.7  & 52.4  & 64.7  & 83.6  & 52.3  & 57.8  & 61.5  & 58.5  & 65.6  & 53.2  & 41.6  & 96.5  & 64.6  & 84.2  & \textbf{65.6}  & 86.5  & 50.7  & 59.4  & 56.5  & 75.2  \\
\Xhline{2\arrayrulewidth}
\textbf{PP1} & \textbf{76.8}  & 72.7  & 54.7  & 85.8  & 78.5  & \textbf{84.5}  & \textbf{74.1}  & \textbf{81.9}  & 90.7  & \textbf{73.5}  & \textbf{77.8}  & \textbf{73.6}  & 64.2  & 62.5  & \textbf{75.0}  & \textbf{65.5}  & 96.6  & 80.3  & 90.9  & 72.1  & 87.5  & 61.2  & \textbf{86.7}  & 71.5  & \textbf{81.4}  \\
\textbf{PP2} & \textbf{54.7}  & $\;\;-$  & 34.8  & $\;\;-$  & $\;\;-$  & \textbf{54.9}  & $\;\;-$  & 60.6  & $\;\;-$  & \textbf{57.0}  & $\;\;-$  & $\;\;-$  & $\;\;-$  & $\;\;-$  & $\;\;-$  & \textbf{56.8}  & $\;\;-$  & 63.0  & $\;\;-$  & 58.4  & $\;\;-$  & 52.9  & \textbf{53.6}  & $\;\;-$  & $\;\;-$  \\
\textbf{PP3} & \textbf{53.4}  & $\;\;-$  & 25.1  & \textbf{61.0}  & $\;\;-$  & \textbf{49.6}  & \textbf{46.1}  & 52.5  & 81.0  & \textbf{48.0}  & \textbf{56.1}  & 60.4  & $\;\;-$  & $\;\;-$  & \textbf{49.1}  & \textbf{46.0}  & $\;\;-$  & 54.3  & $\;\;-$  & 50.7  & $\;\;-$  & 50.6  & \textbf{47.0}  & \textbf{54.7}  & \textbf{75.1}  \\
\hline
\textbf{Avg} & \textbf{68.1}  & 72.7  & 38.2  & \textbf{73.4}  & 78.5  & \textbf{63.0}  & \textbf{60.1}  & 65.0  & 85.8  & \textbf{59.5}  & \textbf{67.0}  & \textbf{67.0}  & 64.2  & 62.5  & \textbf{62.0}  & \textbf{56.1}  & 96.6  & 65.9  & 90.9  & 60.4  & 87.5  & 54.9  & \textbf{62.4}  & \textbf{63.1}  & \textbf{78.2}  \\
\Xhline{2\arrayrulewidth}
\textbf{S1} & 73.9  & \textbf{72.9}  & \textbf{55.9}  & \textbf{86.1}  & \textbf{83.4}  & 83.8  & 72.1  & 73.3  & 90.4  & 60.4  & 70.6  & 71.5  & \textbf{71.6}  & 64.6  & 42.1  & 59.1  & \textbf{97.1}  & 78.6  & 91.6  & 68.7  & 77.0  & \textbf{64.2}  & 83.8  & \textbf{74.4}  & 79.5  \\
\textbf{S2} & 53.3  & $\;\;-$  & \textbf{37.8}  & $\;\;-$  & $\;\;-$  & 53.6  & $\;\;-$  & \textbf{65.3}  & $\;\;-$  & 55.0  & $\;\;-$  & $\;\;-$  & $\;\;-$  & $\;\;-$  & $\;\;-$  & 41.4  & $\;\;-$  & 62.1  & $\;\;-$  & 62.6  & $\;\;-$  & 49.8  & 51.7  & $\;\;-$  & $\;\;-$  \\
\textbf{S3} & 48.0  & $\;\;-$  & 27.2  & 52.8  & $\;\;-$  & 44.7  & 44.2  & 51.1  & 77.2  & 40.7  & 47.5  & 53.7  & $\;\;-$  & $\;\;-$  & 27.3  & 35.7  & $\;\;-$  & 54.4  & $\;\;-$  & 52.4  & $\;\;-$  & \textbf{53.1}  & 43.3  & 48.0  & 62.3  \\
\hline
\textbf{Avg} & 65.1  & \textbf{72.9}  & 40.3  & 69.4  & \textbf{83.4}  & 60.7  & 58.1  & 63.2  & 83.8  & 52.0  & 59.0  & 62.6  & \textbf{71.6}  & 64.6  & 34.7  & 45.4  & \textbf{97.1}  & 65.0  & 91.6  & 61.2  & 77.0  & \textbf{55.7}  & 59.6  & 61.2  & 70.9  \\
\Xhline{2\arrayrulewidth}
\textbf{C1} & 75.5  & 72.0  & 55.3  & 83.6  & 75.0  & 83.9  & 65.6  & 81.8  & \textbf{91.9}  & 68.1  & 74.5  & 71.1  & 66.8  & \textbf{70.4}  & 68.1  & 55.6  & \textbf{97.1}  & \textbf{83.1}  & \textbf{92.7}  & 78.9  & \textbf{92.6}  & 58.8  & 85.5  & 67.7  & 71.8  \\
\textbf{C2} & 52.1  & $\;\;-$  & 36.6  & $\;\;-$  & $\;\;-$  & 52.9  & $\;\;-$  & 63.4  & $\;\;-$  & 54.9  & $\;\;-$  & $\;\;-$  & $\;\;-$  & $\;\;-$  & $\;\;-$  & 42.4  & $\;\;-$  & 64.1  & $\;\;-$  & 57.7  & $\;\;-$  & \textbf{54.4}  & 42.7  & $\;\;-$  & $\;\;-$  \\
\textbf{C3} & 49.6  & $\;\;-$  & \textbf{29.1}  & 58.7  & $\;\;-$  & 47.7  & 36.2  & \textbf{55.3}  & \textbf{81.5}  & 40.4  & 55.8  & \textbf{60.7}  & $\;\;-$  & $\;\;-$  & 26.4  & 34.4  & $\;\;-$  & \textbf{58.7}  & $\;\;-$  & 50.8  & $\;\;-$  & 52.3  & 37.4  & 50.8  & 67.0  \\
\hline
\textbf{Avg} & 66.3  & 72.0  & \textbf{40.3}  & 71.2  & 75.0  & 61.5  & 50.9  & \textbf{66.8}  & \textbf{86.7}  & 54.5  & 65.2  & 65.9  & 66.8  & \textbf{70.4}  & 47.2  & 44.1  & \textbf{97.1}  & \textbf{68.6}  & \textbf{92.7}  & 62.5  & \textbf{92.6}  & 55.2  & 55.2  & 59.2  & 69.4  \\
     \Xhline{2\arrayrulewidth}
    \end{tabular}
    \caption{\textbf{Fine-grained semantic segmentation results (shape mIoU \%).} Algorithm \textbf{P}, \textbf{P\textsuperscript{+}}, \textbf{S} and \textbf{C} refer to PointNet \cite{qi2017pointnet}, PointNet++ \cite{qi2017pointnet++}, SpiderCNN \cite{xu2018spidercnn} and PointCNN \cite{li2018pointcnn}, respectively. The number \textbf{1}, \textbf{2} and \textbf{3} refer to the three levels of segmentation: coarse-, middle- and fine-grained. We put short lines for the levels that are not defined.}
    \label{tab:sem_seg_per_shape_result}
\end{table*}

Table~\ref{tab:confusion_scores} compares the annotation consistency before and after the template refinement process. We compute the confusion matrices at the most fine-grained segmentation level. After the template refinement, the data consistency score is 83.3\% on average, having 13.5\% improvement over the raw annotation. The template refinement process improves the annotation consistency by a clear margin. This also reflects the complexity of the task in terms of annotating fine-grained part concepts. Future works may investigate how to further design better templates with less part ambiguities.

\subsection{More Visualization of Hierarchical Templates}
Figure~\ref{fig:template_visu1}, \ref{fig:template_visu2} and \ref{fig:template_visu3} show more visualization for the expert-designed hierarchical templates after resolving the data inconsistency and conducting template refinements. We show the lamp template in the main paper.

%% file: tex_supp/tasks.tex
In this section, we provide more details about the architectures and training details for the benchmark algorithms. We also present additional evaluation metrics, shape mean Intersection-over-Union (shape mIoU) and shape mean Average-Precision (Shape mAP), and report the quantitative results using these metrics.

\subsection{Fine-grained Semantic Segmentation}
\label{sec:task1}

\paragraph{More Architecture and Training Details}
We follow the default architectures and training hyper-parameters used in the original papers: PointNet~\cite{qi2017pointnet}, PointNet++~\cite{qi2017pointnet++}, SpiderCNN~\cite{xu2018spidercnn} and PointCNN~\cite{li2018pointcnn}, except the following few modifications:
\begin{itemize}
    \item Instead of training one network for all object categories as done in the four prior works, we train separate networks for each object category at each segmentation level. This is mainly to handle the increase in the number of parts for fine-grained part segmentation. Originally, there are only 50 parts for all 16 object categories using the coarse ShapeNet Part dataset~\cite{yi2016scalable}. Now, using PartNet, there could be 480 different part semantics in total. Also, due to the data imbalance among different object categories, training a single network may overfit to the big categories.
    \item We change the input point cloud size to 10,000. The original papers usually sample 1,000, 2,000 or 4,000 points and input to the networks. PartNet suggests to use at least 10,000 to guarantee enough point sampling over small fine-grained parts, \eg a door handle, or a small button.
    \item We reduce the batch sizes for training the networks if necessary. Since we use point cloud size 10,000, to fit the training in NVIDIA TITAN XP GPU 12G memory, we need to adjust the training batch size accordingly. For PointNet~\cite{qi2017pointnet}, PointNet++~\cite{qi2017pointnet++}, SpiderCNN~\cite{xu2018spidercnn} and PointCNN~\cite{li2018pointcnn}, we use batch size of 24, 24, 2 and 4 respectively.
    \item We only input 3D coordinates as inputs to all the networks for fair comparison. Although the 3D CAD models in ShapeNet~\cite{chang2015shapenet} usually provide additional features, \eg opacity, point normals, textures and material information, there is no guarantee for the quality of such information. Thus, we choose not to use them as the inputs. Also, only using pure geometry potentially increase the network generalizability to unseen objects or real scans~\cite{qi2017pointnet}. PointNet++~\cite{qi2017pointnet++} and SpiderCNN~\cite{xu2018spidercnn} by defaults take advantage of the point normals as additional inputs. In this paper, we remove such inputs to the networks. However, point normals can be estimated from the point clouds. We leave this as a future work.
\end{itemize}

\begin{table*}[t]
    \footnotesize
    \centering
     \begin{tabular}[width=\linewidth]{@{\hspace{0.15cm}}p{1.4cm}|@{\hspace{0.15cm}}p{0.4cm}|@{\hspace{0.15cm}}p{0.45cm}p{0.45cm}p{0.45cm}p{0.45cm} p{0.45cm}p{0.45cm}p{0.45cm}p{0.45cm} p{0.45cm}p{0.45cm}p{0.45cm}p{0.45cm} p{0.45cm}p{0.45cm}p{0.45cm}p{0.45cm} p{0.5cm}}
    \Xhline{2\arrayrulewidth}
     & \textbf{Avg} & \textbf{Bed} & \textbf{Bott} & \textbf{Chair} & \textbf{Clock} & \textbf{Dish} & \textbf{Disp} & \textbf{Door} & \textbf{Ear} & \textbf{Fauc} & \textbf{Knife} & \textbf{Lamp} & \textbf{Micro} & \textbf{Frid} & \textbf{Stora} & \textbf{Table} & \textbf{Trash} & \textbf{Vase} \\ \Xhline{2\arrayrulewidth}
     \textbf{Bottom-Up} & 65.9  & 42.0  & 74.3  & 63.8  & 64.1  & \textbf{66.3}  & 84.2  & 61.4  & 70.0  & \textbf{74.2}  & 67.1  & 62.7  & \textbf{63.0}  & 60.8  & 57.8  & 65.7  & 62.8  & 80.9 \\
     \textbf{Top-Down} & 65.9  & 42.0  & 73.7  & 62.3  & \textbf{65.5}  & 64.0  & 85.5  & 63.1  & 71.1  & 73.5  & \textbf{68.8}  & 63.3  & 62.7  & 58.8  & 57.6  & 66.2  & 63.0  & 79.3 \\
     \textbf{Ensemble} & \textbf{66.6}  & \textbf{42.9}  & \textbf{74.4}  & \textbf{64.3}  & \textbf{65.5}  & 62.7  & \textbf{85.8}  & \textbf{63.7}  & \textbf{71.7}  & 74.0  & 66.7  & \textbf{63.4}  & 61.9  & \textbf{61.5}  & \textbf{60.6}  & \textbf{67.5}  & \textbf{64.0}  & \textbf{82.2} \\
     \Xhline{2\arrayrulewidth}
    \end{tabular}
    \caption{\textbf{Hierarchical segmentation results (shape mIoU \%).} We present the hierarchical segmentation performances for three baseline methods: bottom-up, top-down and ensemble. We conduct experiments on 17 out of 24 categories with tree depth bigger than 1.}
    \label{tab:hier_seg_per_part_category_result}
\end{table*}

\paragraph{Shape mIoU Metric and Results}
\label{sec:shape_miou}
We introduce the shape mean Intersect-over-Union (Shape mIoU) evaluation metric as a secondary metric to the Part-category mIoU metric in the main paper. Shape mIoU metric considers shapes as evaluation units and measures how an algorithm segment an average shape in the object category. In contrast, Part-category mIoU reports the average performance over all part semantics and indicates how an algorithm performs for any given part category.

Shape mIoU is widely used on ShapeNet Part dataset~\cite{yi2016scalable} for 3D shape coarse semantic segmentation~\cite{qi2017pointnet, qi2017pointnet++, xu2018spidercnn, li2018pointcnn}. We propose a slightly different version for fine-grained semantic segmentation. For each test shape, we first compute the IoU for each part semantics that either presents in the ground-truth or is predicted by the algorithm, and then we calculate the mean IoU for this shape. We remove the ground-truth unlabeled points from the evaluation. Finally, we calculate the Shape mIoU by averaging mIoU over all test shape instances. 

We benchmark the four algorithms using Shape mIoU in Table~\ref{tab:sem_seg_per_shape_result}. Besides the Shape mIoU scores for each object category at each segmentation level, we also report the average across levels for each object categories and further calculate the average over all object categories.

We observe that PointNet++~\cite{qi2017pointnet++} achieves the best performance using the Shape mIoU metric, while PointCNN~\cite{li2018pointcnn} performs the best using the Part-category mIoU metric. The Part-category mIoU metric considers all part semantics equally while the Shape mIoU metric considers all shapes equally. We observe an unbalanced counts for different part semantics in most object categories, \eg there are much more chair legs than chair wheels. To achieve good numbers on Part-category mIoU, a segmentation algorithm needs to perform equally well on both frequent parts and rare parts, while the Shape mIoU metric bias over the frequently observed parts.

\subsection{Hierarchical Semantic Segmentation}
\label{sec:task2}

We describe the architecture and training details for the three baseline methods we propose for hierarchical semantic segmentation in the main paper. All three methods use PointNet++~\cite{qi2017pointnet++} segmentation network as the network backbone. The difference of the three methods is mainly on the training and inference strategies to enforce the tree knowledge to the final prediction.

\paragraph{The Bottom-up Method}
The bottom-up method learns a network to perform segmentation at the most fine-grained leaf part semantics. We use the PointNet++~\cite{qi2017pointnet++} segmentation network with a softmax activation layer as the network architecture. At inference time, we use the ground-truth tree hierarchy to gather the prediction for the parent nodes. The parent node prediction is the sum of all its children node predictions. Even though we only train for the leaf node parts, the parent history is implicitly encoded. For example, we define vertical bars for both chair back and chair arm, but they are two different leaf node parts: \textit{chair back vertical bar} and \textit{chair arm vertical bar}. 

In the ground-truth annotation, all the points in the point cloud belong to the root node. Each point is assigned a path of labels from the root node to some intermediate node in the tree. The paths for most points are all the way down to the leaf levels while some points may not. For example, a point on a bed blanket (removed from evaluation since it cannot be distinguished without color information) may be assigned with labels \{\textit{bed}, \textit{bed unit}, \textit{sleeping area}\} in the ground-truth annotation. The part \textit{sleeping area} is not a leaf part. For such cases, we introduce an additional leaf node \textit{other} for each parent node in the tree and consider them in the training.

\paragraph{The Top-down Method}
The top-down method learns a multi-labeling task for all the part semantics in the tree, considering both the leaf nodes and the parent nodes. Compared to the bottom-up method, the top-down method takes advantage of the tree structures at training time.

Assuming there are $T$ tree nodes in the hierarchy, we train a PointNet++~\cite{qi2017pointnet++} segmentation network for a $T$-way classification for each point. We apply a softmax activation layer to enforce label mutual exclusiveness. For a point with the ground-truth labels $\{y_1, y_2, y_3\}$ and prediction softmax scores $\{s_i|i=1,2,\cdots, T\}$, we train the labels using a multi-labeling cross-entropy loss
\begin{equation}
    L=-\log(s_{y_1})-\log(s_{y_2})-\log(s_{y_3})
\end{equation}
to increase the values of all the three label predictions over the rest labels.

\begin{table*}[th!]
    \footnotesize
    \centering
    \begin{tabular}[width=\linewidth]{@{\hspace{0.1cm}}p{0.3cm}|@{\hspace{0.1cm}}p{0.35cm}|@{\hspace{0.1cm}}p{0.15cm}p{0.15cm}p{0.2cm}p{0.25cm} p{0.35cm}p{0.35cm}p{0.25cm}p{0.25cm} p{0.25cm}p{0.15cm}p{0.25cm}p{0.15cm} p{0.15cm}p{0.35cm}p{0.35cm}p{0.15cm} p{0.35cm}p{0.25cm}p{0.25cm}p{0.15cm} p{0.35cm}p{0.35cm}p{0.35cm}p{0.35cm}}
    \Xhline{2\arrayrulewidth}
     & \textbf{Avg} & \textbf{Bag} & \textbf{Bed} & \textbf{Bott} & \textbf{Bowl} & \textbf{Chair} & \textbf{Clock} & \textbf{Dish} & \textbf{Disp} & \textbf{Door} & \textbf{Ear} & \textbf{Fauc} & \textbf{Hat} & \textbf{Key} & \textbf{Knife} & \textbf{Lamp} & \textbf{Lap} & \textbf{Micro} & \textbf{Mug} & \textbf{Frid} & \textbf{Scis} & \textbf{Stora} & \textbf{Table} & \textbf{Trash} & \textbf{Vase} \\ 
     \Xhline{2\arrayrulewidth}
     \textbf{S1} & 72.5  & 62.8  & 38.7  & 76.7  & 83.2  & 91.5  & 41.5  & \textbf{81.4}  & 91.3  & 71.2  & 81.4  & 82.2  & 71.9  & 23.2  & \textbf{78.0}  & 60.3  & \textbf{100}  & 76.2  & 94.3  & 60.6  & 74.9  & 55.0  & 80.1  & 76.1  & 87.1  \\
\textbf{S2} & 50.2  & $\;\;-$  & 22.7  & $\;\;-$  & $\;\;-$  & 51.1  & $\;\;-$  & \textbf{78.7}  & $\;\;-$  & 43.3  & $\;\;-$  & $\;\;-$  & $\;\;-$  & $\;\;-$  & $\;\;-$  & 49.1  & $\;\;-$  & 68.6  & $\;\;-$  & 42.9  & $\;\;-$  & 51.9  & 43.7  & $\;\;-$  & $\;\;-$  \\
\textbf{S3} & 50.2  & $\;\;-$  & 17.5  & 66.5  & $\;\;-$  & 42.3  & 40.7  & 59.3  & 83.9  & 29.0  & 60.2  & 61.6  & $\;\;-$  & $\;\;-$  & 55.0  & 37.6  & $\;\;-$  & 53.7  & $\;\;-$  & 30.6  & $\;\;-$  & 45.1  & 37.8  & 50.0  & 82.0  \\
\hline
\textbf{Avg} & 64.2  & 62.8  & 26.3  & 71.6  & 83.2  & 61.6  & 41.1  & 73.1  & 87.6  & 47.8  & 70.8  & 71.9  & 71.9  & 23.2  & 66.5  & 49.0  & \textbf{100}  & 66.2  & 94.3  & 44.7  & 74.9  & 50.7  & 53.8  & 63.0  & 84.6  \\
\Xhline{2\arrayrulewidth}
\textbf{O1} & \textbf{80.3}  & \textbf{78.4}  & \textbf{62.2}  & \textbf{80.8}  & \textbf{83.8}  & \textbf{94.9}  & \textbf{74.6}  & \textbf{81.4}  & \textbf{94.3}  & \textbf{76.1}  & \textbf{87.1}  & \textbf{86.5}  & \textbf{77.8}  & \textbf{44.5}  & 76.6  & \textbf{65.0}  & \textbf{100}  & \textbf{79.5}  & \textbf{95.3}  & \textbf{79.0}  & \textbf{87.6}  & \textbf{62.7}  & \textbf{88.1}  & \textbf{82.3}  & \textbf{89.0}  \\
\textbf{O2} & \textbf{60.5}  & $\;\;-$  & \textbf{29.4}  & $\;\;-$  & $\;\;-$  & \textbf{64.7}  & $\;\;-$  & 75.4  & $\;\;-$  & \textbf{61.1}  & $\;\;-$  & $\;\;-$  & $\;\;-$  & $\;\;-$  & $\;\;-$  & \textbf{56.8}  & $\;\;-$  & \textbf{78.2}  & $\;\;-$  & \textbf{61.7}  & $\;\;-$  & \textbf{57.4}  & \textbf{59.4}  & $\;\;-$  & $\;\;-$  \\
\textbf{O3} & \textbf{57.7}  & $\;\;-$  & \textbf{22.1}  & \textbf{68.3}  & $\;\;-$  & \textbf{58.4}  & \textbf{53.7}  & \textbf{67.5}  & \textbf{84.8}  & \textbf{38.0}  & \textbf{62.4}  & \textbf{66.8}  & $\;\;-$  & $\;\;-$  & \textbf{63.5}  & \textbf{45.8}  & $\;\;-$  & \textbf{54.0}  & $\;\;-$  & \textbf{45.0}  & $\;\;-$  & \textbf{52.6}  & \textbf{52.5}  & \textbf{58.7}  & \textbf{86.4}  \\
\hline
\textbf{Avg} & \textbf{72.2}  & \textbf{78.4}  & \textbf{37.9}  & \textbf{74.6}  & \textbf{83.8}  & \textbf{72.7}  & \textbf{64.2}  & \textbf{74.8}  & \textbf{89.5}  & \textbf{58.4}  & \textbf{74.8}  & \textbf{76.6}  & \textbf{77.8}  & \textbf{44.5}  & \textbf{70.1}  & \textbf{55.8}  & \textbf{100}  & \textbf{70.6}  & \textbf{95.3}  & \textbf{61.9}  & \textbf{87.6}  & \textbf{57.6}  & \textbf{66.7}  & \textbf{70.5}  & \textbf{87.7}  \\
     \Xhline{2\arrayrulewidth}
    \end{tabular}
    \caption{\textbf{Instance segmentation results (shape mAP \%, IoU threshold 0.5).} Algorithm \textbf{S} and \textbf{O} refer to SGPN~\cite{wang2018sgpn} and our proposed method respectively. The number \textbf{1}, \textbf{2} and \textbf{3} refer to the three levels of segmentation: coarse-, middle- and fine-grained. We put short lines for the levels that are not defined.}
    \label{tab:ins_seg_per_part_category_result}
\end{table*}

\paragraph{The Ensemble Method}
The ensemble method trains multiple neural networks at different levels of segmentation as defined in the fine-grained semantic segmentation task. The key idea is that conducting segmentation at the coarse-, middle- and fine-grained levels separately may learn different features that work the best at each level. Compared to the bottom-up method that we only train at the most fine-grained level, additional signal at the coarse level helps distinguish the coarse-level part semantics more easily. For example, the local geometric features for both chair back vertical bars and chair arm vertical bars may be very similar, but the coarse-level semantics may distinguish chair backs and chair arms better.

During the training, we train 2$\sim$3 networks at multiple levels of segmentation. At the inference time, we perform a joint inference considering the prediction scores from all the networks. We use a path-voting strategy: for each path from the root node to the leaf node, we calculate the average log-likelihood over the network prediction scores after applying the softmax activations, and select the path with the highest score as the joint label predictions.

\paragraph{Shape mIoU Metric and Results}
Similar to Sec~\ref{sec:shape_miou}, we define Shape mIoU for hierarchical segmentation. The mIoU for each shape is calculated over the part semantics in the entire hierarchical template that are either predicted by the network or included in the ground-truth. The unrelated parts are not taken into consideration.
Table~\ref{tab:hier_seg_per_part_category_result} shows the quantitative evaluation for the three baseline methods. We observe similar performance for the three methods, with the ensemble method works slightly better.

\subsection{Instance Segmentation}
\label{sec:task3}

\paragraph{More Architecture and Training Details}
To train our proposed method, we use batch size 32, learning rate 0.001, and the default batch normalization settings used in the PointNet++~\cite{qi2017pointnet++}. 

For SGPN~\cite{wang2018sgpn}, we use two-stage training as suggested by the authors of~\cite{wang2018sgpn}. We first pretrain the PointNet++ semantic segmentation branch using batch size 32 and learning rate 0.001, with the default batch normalization as in PointNet++. And then, we jointly train for the semantic segmentation, similarity score matrix and confidence scores with batch size 1 and learning rate 0.0001. As suggested in the original SGPN paper, for the first five epochs of the joint training, we only turn on the loss for training the similarity scores matrix. The rest training epochs are done with the full losses switched on. We have to use batch size 1 because the input point cloud has the size of 10,000 and thus the similarity score matrix forms a $10,000\times 10,000$ matrix, which occupies too much GPU memory. Our proposed method is more memory-efficient, compared to SGPN. We also observe that our training is much faster than SPGN. We train all the networks until convergence.

\paragraph{Shape mAP Metric and Results}
We define Shape mean Average-Precision (Shape mAP) metric as a secondary metric to the Part-category mAP metric in the main paper. Similar to the Shape mIoU scores we use in Sec~\ref{sec:task1}~and~\ref{sec:task2}, Shape mAP reports the part instance segmentation performance on an average shape in a object category. It averages across the test shapes, instead of averaging across all different part semantics, as benchmarked by Part-category mAP in the main paper.

To calculate Shape mAP for a test shape, we consider the AP for the part semantics that occur either in the ground-truth or the prediction for the given shape and compute their average as the mean AP score. Then, we average the mAP across all test shapes within a object category. Table~\ref{tab:ins_seg_per_part_category_result} reports the part instance segmentation performance under the Shape mAP scores. We see a clear performance improvement of the proposed method over SGPN.

%% file: tex_supp/figures-min.tex
\begin{figure*}[t!]
\centering
\includegraphics[width=0.9\linewidth]{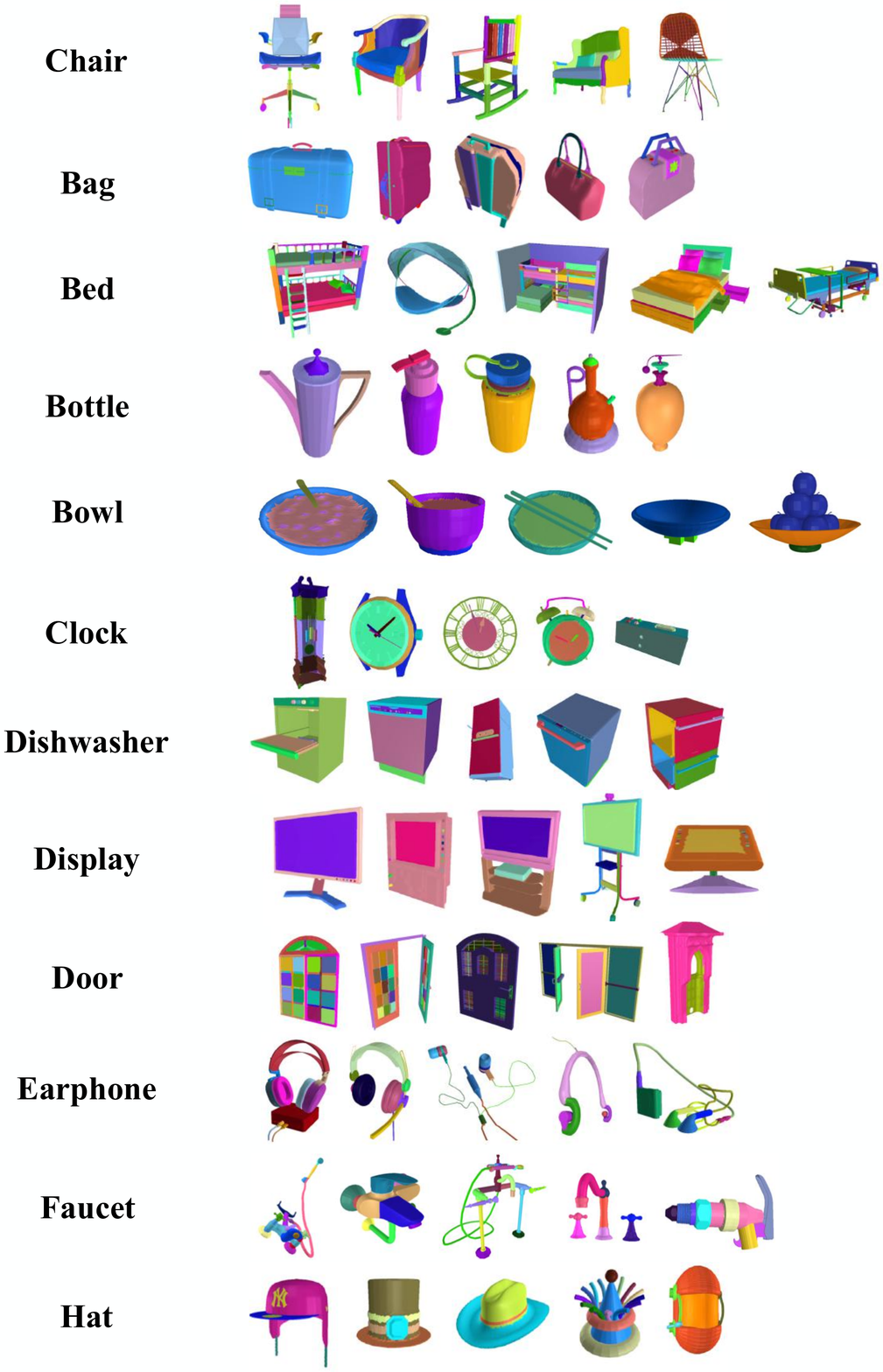}
\caption{\textbf{Fine-grained instance-level segmentation visualization (1/2).} We present visualization for example fine-grained instance-level segmentation annotations for chair, bag, bed, bottle, bowl, clock, dishwasher, display, door, earphone, faucet, and hat.}
\label{fig:fine_grained_visu1}
\end{figure*}

\begin{figure*}[t!]
\centering
\includegraphics[width=0.9\linewidth]{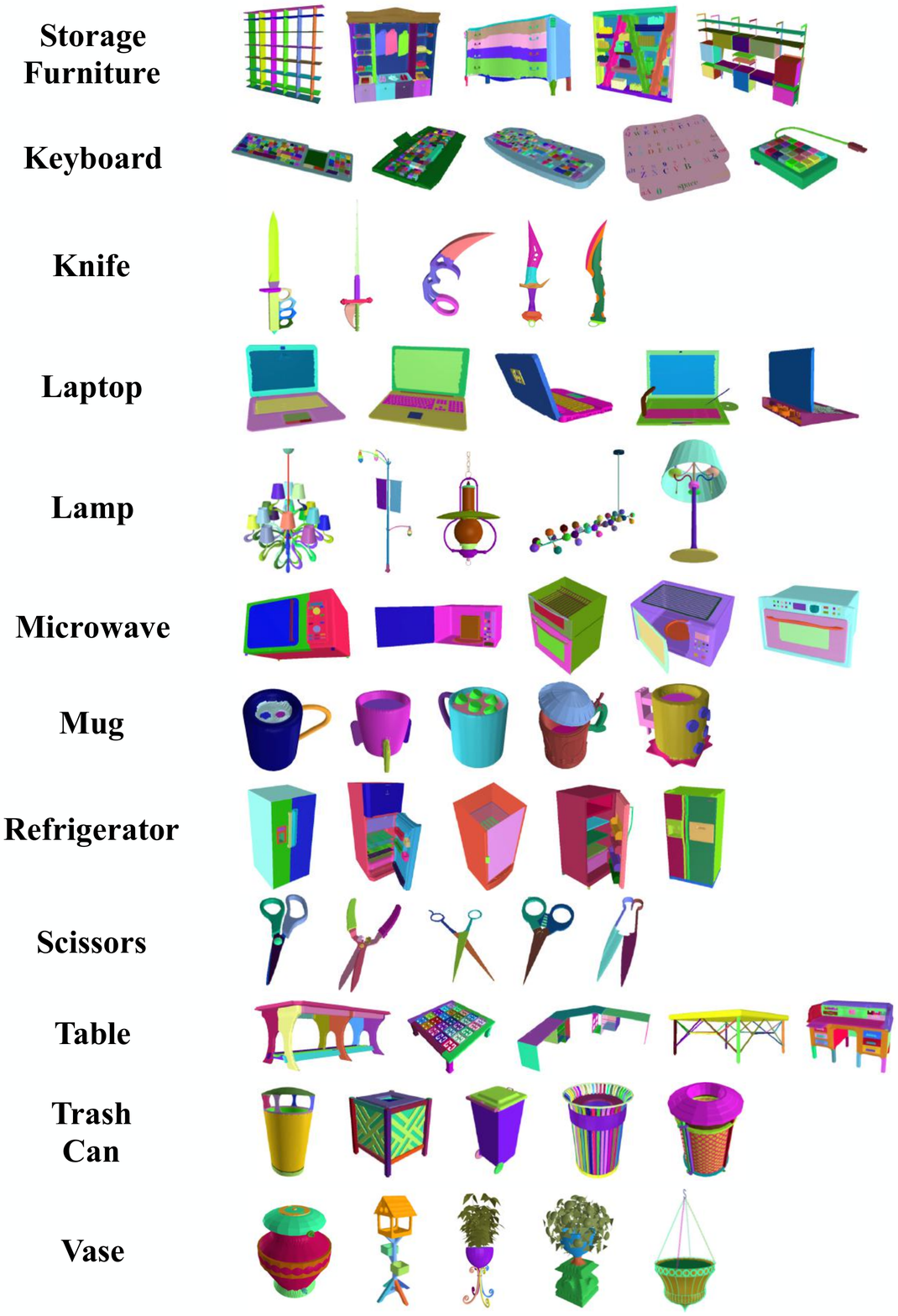}
\caption{\textbf{Fine-grained instance-level segmentation visualization (2/2).} We present visualization for example fine-grained instance-level segmentation annotations for storage furniture, keyboard, knife, laptop, lamp, microwave, mug, refrigerator, scissors, table, trash can, and vase.}
\label{fig:fine_grained_visu2}
\end{figure*}

\begin{figure*}[t!]
\centering
\includegraphics[width=0.98\linewidth]{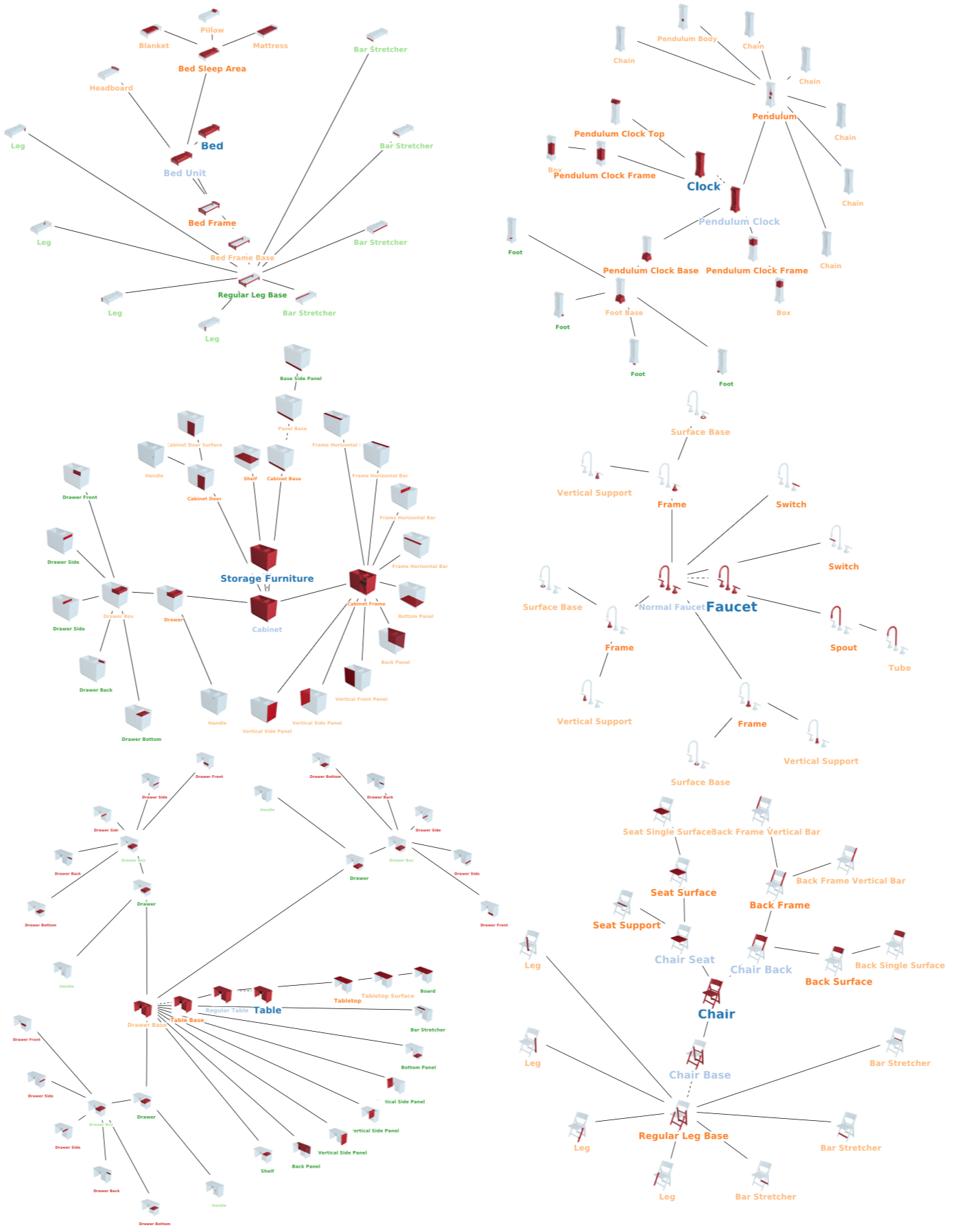}
\caption{\textbf{Hierarchical instance-level segmentation visualization (1/3).} We present visualization for example hierarchical instance-level segmentation annotations for bed, clock, storage furniture, faucet, table, and chair. The lamp examples are shown in the main paper. The And-nodes are drawn in solid lines and Or-nodes in dash lines.}
\label{fig:hier_visu1}
\end{figure*}

\begin{figure*}[t!]
\centering
\includegraphics[width=0.98\linewidth]{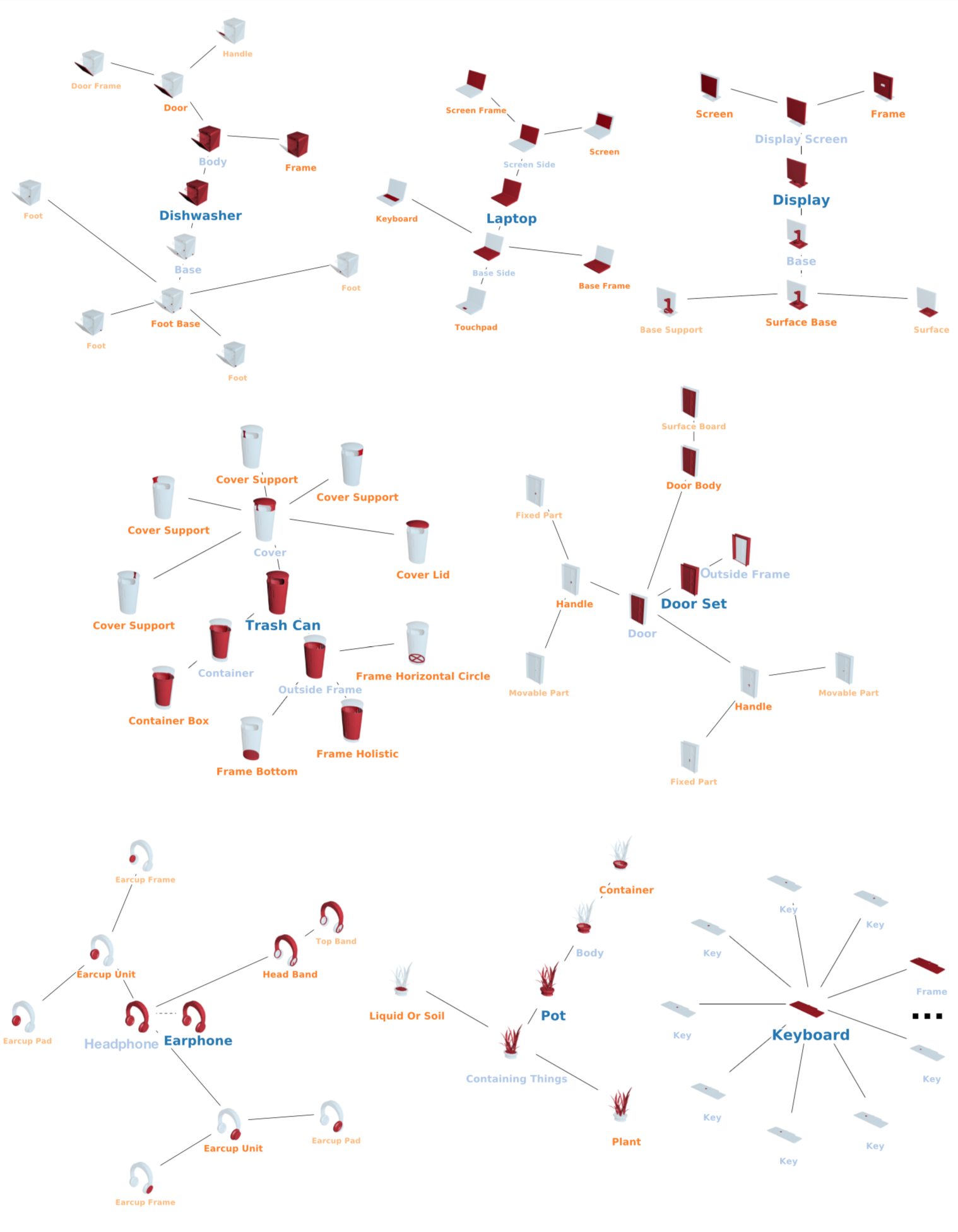}
\caption{\textbf{Hierarchical instance-level segmentation visualization (2/3).} We present visualization for example hierarchical instance-level segmentation annotations for dishwasher, laptop, display, trash can, door (door set), earphone, vase (pot), and keyboard. The lamp examples are shown in the main paper. The And-nodes are drawn in solid lines and Or-nodes in dash lines.}
\label{fig:hier_visu2}
\end{figure*}

\begin{figure*}[t!]
\centering
\includegraphics[width=0.98\linewidth]{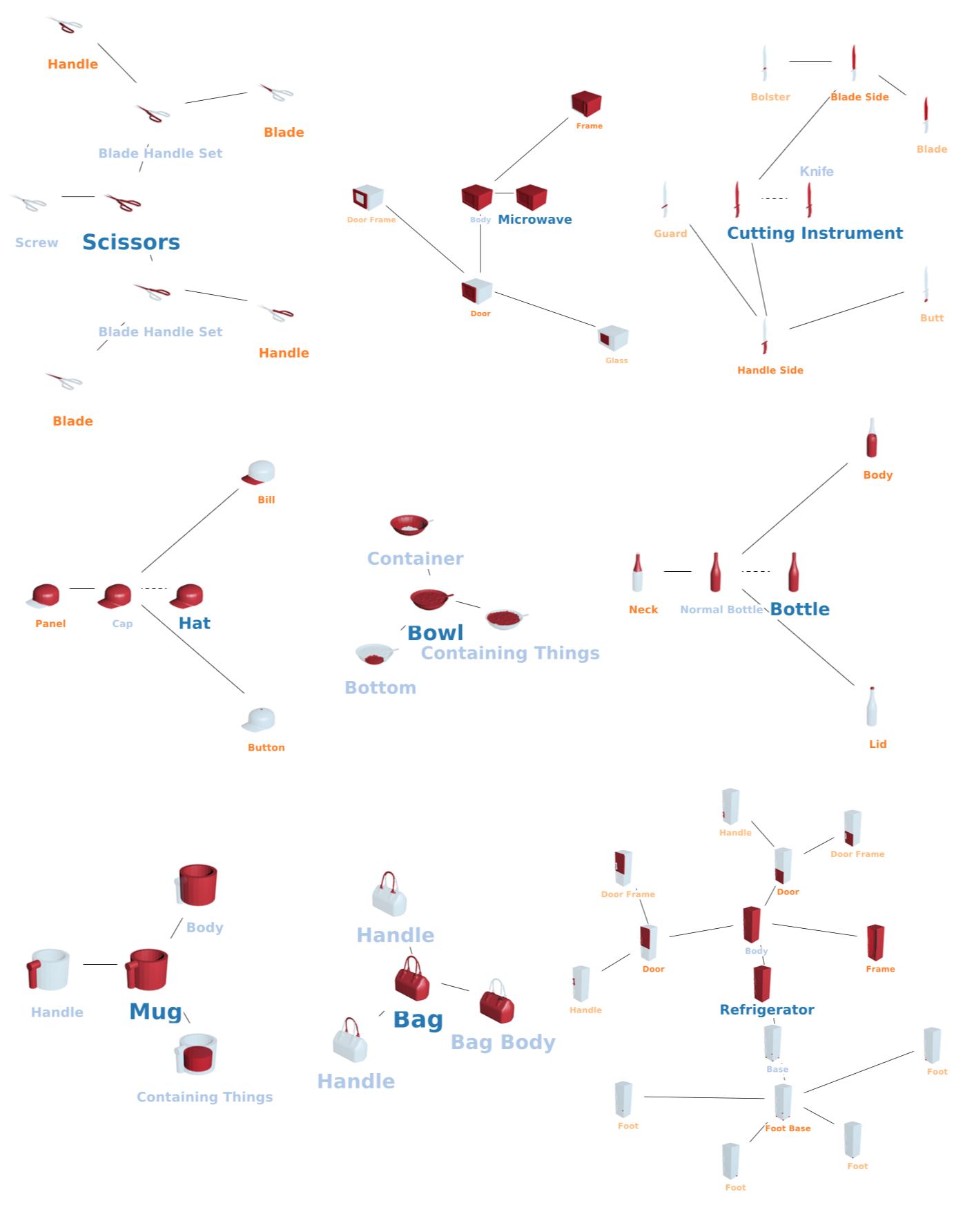}
\caption{\textbf{Hierarchical instance-level segmentation visualization (3/3).} We present visualization for example hierarchical instance-level segmentation annotations for scissors, microwave, knife (cutting instrument), hat, bowl, bottle, mug, bag, and refrigerator. The lamp examples are shown in the main paper. The And-nodes are drawn in solid lines and Or-nodes in dash lines.}
\label{fig:hier_visu3}
\end{figure*}

\begin{figure*}[t!]
\centering
\includegraphics[width=0.98\linewidth]{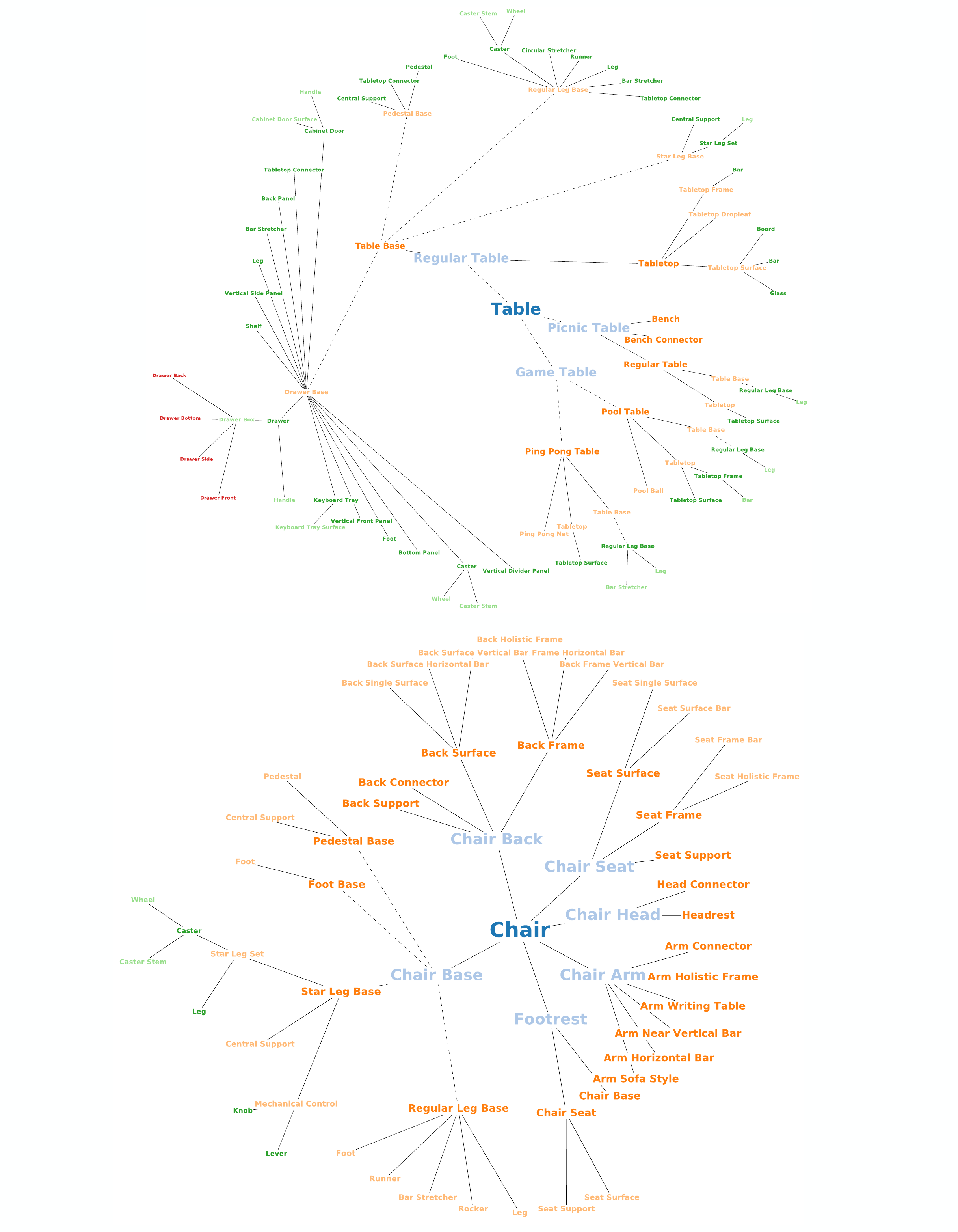}
\caption{\textbf{Template visualization (1/3).} We present the templates for table and chair. The lamp template is shown in the main paper. The And-nodes are drawn in solid lines and Or-nodes in dash lines.}
\label{fig:template_visu1}
\end{figure*}

\begin{figure*}[t!]
\centering
\includegraphics[width=0.98\linewidth]{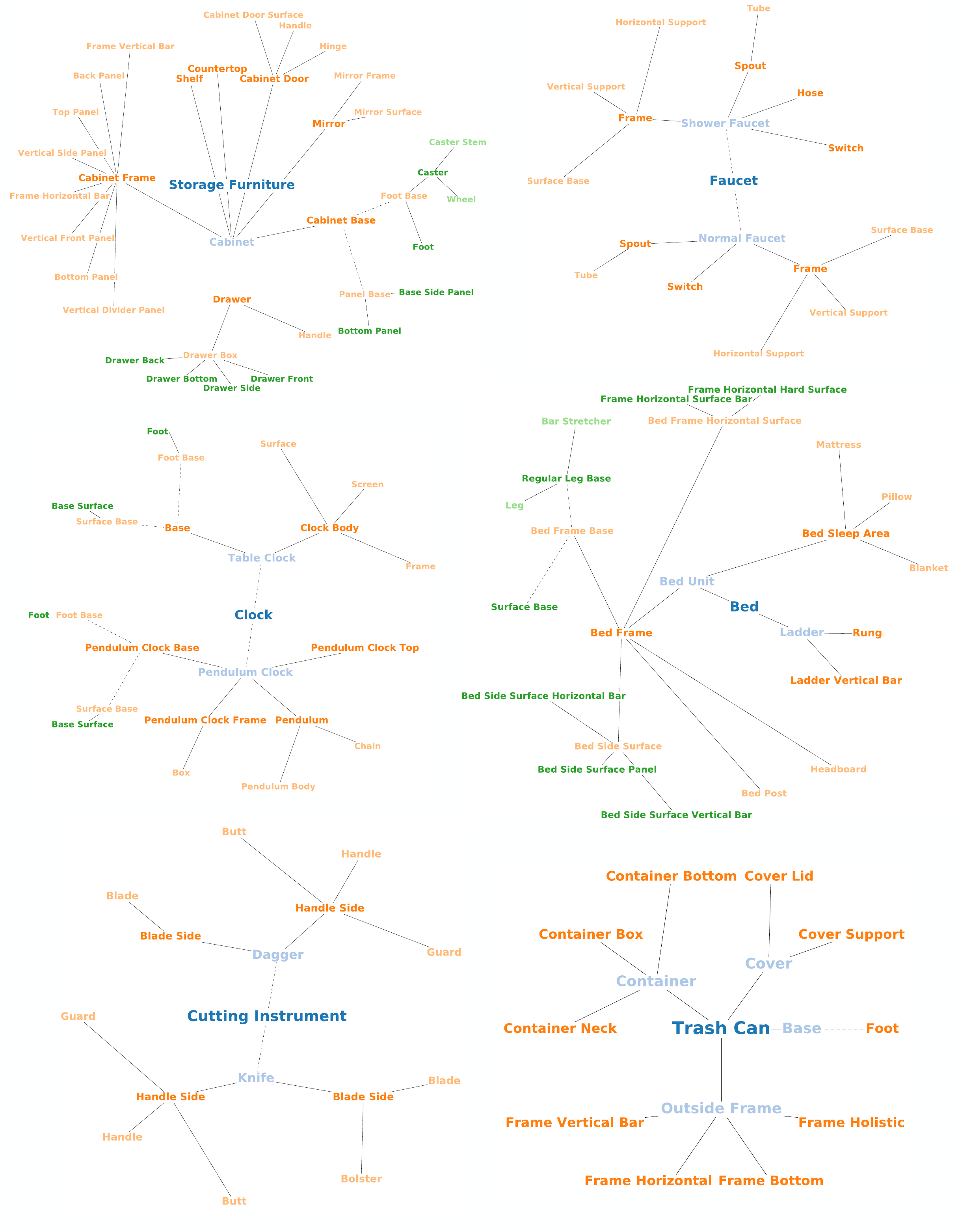}
\caption{\textbf{Template visualization (2/3).} We present the templates for storage furniture, faucet, clock, bed, knife (cutting instrument), and trash can. The lamp template is shown in the main paper. The And-nodes are drawn in solid lines and Or-nodes in dash lines.}
\label{fig:template_visu2}
\end{figure*}

\begin{figure*}[t!]
\centering
\includegraphics[width=0.98\linewidth]{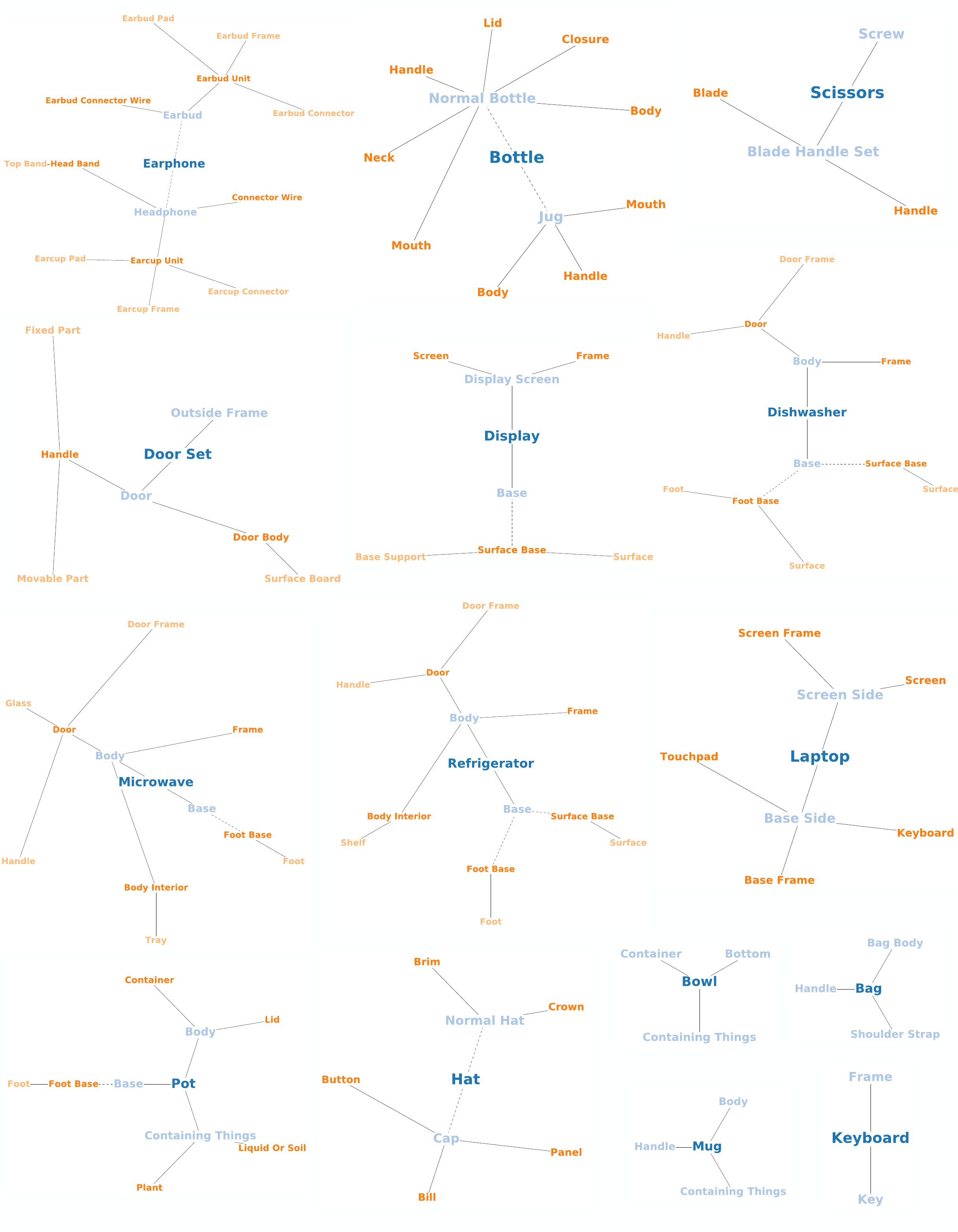}
\caption{\textbf{Template visualization (3/3).} We present the templates for earphone, bottle, scissors, door (door set), display, dishwasher, microwave, refrigerator, laptop, vase (pot), hat, bowl, bag, mug, and keyboard. The lamp template is shown in the main paper. The And-nodes are drawn in solid lines and Or-nodes in dash lines.}
\label{fig:template_visu3}
\end{figure*}